\documentclass[journal]{IEEEtran}
\ifCLASSINFOpdf
  \usepackage[pdftex]{graphicx}
\else
  \usepackage[dvips]{graphicx}
\fi
\usepackage{amsmath}
\usepackage{amssymb}
\usepackage{amsfonts}
\usepackage{multirow}
\usepackage{array}
\usepackage{url}

\begin{document}
%
\title{Random VLAD based Deep Hashing for Efficient Image Retrieval}
%
%
%

\author{Li~Weng,~\IEEEmembership{}
        Lingzhi~Ye,~\IEEEmembership{}        
        Jiangmin~Tian,~\IEEEmembership{}      
        Jiuwen~Cao,~\IEEEmembership{}
        and~Jianzhong~Wang~\IEEEmembership{}
\thanks{The authors are with the School of Automation (Artificial Intelligence), Hangzhou Dianzi University, Hangzhou, 310018 China. Corresponding author: Li Weng, e-mail: lweng@hdu.edu.cn.}
}

\maketitle

\begin{abstract}
Image hash algorithms generate compact binary representations that can be quickly matched by Hamming distance, thus become an efficient solution for large-scale image retrieval.  This paper proposes RV-SSDH, a deep image hash algorithm that incorporates the classical VLAD (vector of locally aggregated descriptors) architecture into neural networks. Specifically, a novel neural network component is formed by coupling a random VLAD layer with a latent hash layer through a transform layer. This component can be combined with convolutional layers to realize a hash algorithm. We implement RV-SSDH as a point-wise algorithm that can be efficiently trained by minimizing classification error and quantization loss. Comprehensive experiments show this new architecture significantly outperforms baselines such as NetVLAD and SSDH, and offers a cost-effective trade-off in the state-of-the-art. In addition, the proposed random VLAD layer leads to satisfactory accuracy with low complexity, thus shows promising potentials as an alternative to NetVLAD.
\end{abstract}


%
\IEEEpeerreviewmaketitle

\section{Introduction}
%
%
%
%

\IEEEPARstart{C}{ontent}-based image search is one of the basic challenges in the management of massive multimedia data. Solutions typically rely on finding appropriate features which are robust and discriminative. 
Previously, numerous feature detectors and descriptors have been proposed, such as GIST~\cite{Oliva2001}, SIFT~\cite{Lowe2004}, BRIEF~\cite{Calonder2010}, etc.
In the big data era, the amount of visual content is drastically increasing, and efficient image search has become a more demanding task. In order to have more \emph{compact} feature representations, two approaches have been developed, which can be summarized as ``aggregation'' and ``quantization''. The former transforms multiple descriptors into a concise form; the latter reduces the precision of a descriptor. 
Typical examples include the bag-of-features (BoF) representation~\cite{Sivic2003} and the product quantization (PQ)~\cite{Jegou2011}. Along these lines of work, this paper focuses on two subsequent techniques. The first is VLAD (vector of locally aggregated descriptors)~\cite{Jegou2010}. This representation is related to BoF but carries more discriminative information. 
The other technique is called hashing~\cite{Wang2016}, an active domain in recent years. It generally means to convert features into binary vectors, called \emph{hash values}. The comparison of hash values can be extremely fast, because Hamming distance is used, which can be efficiently computed by the Exclusive Or (XOR) operation.

\begin{figure}[t]
\centering
\includegraphics[scale=0.3]{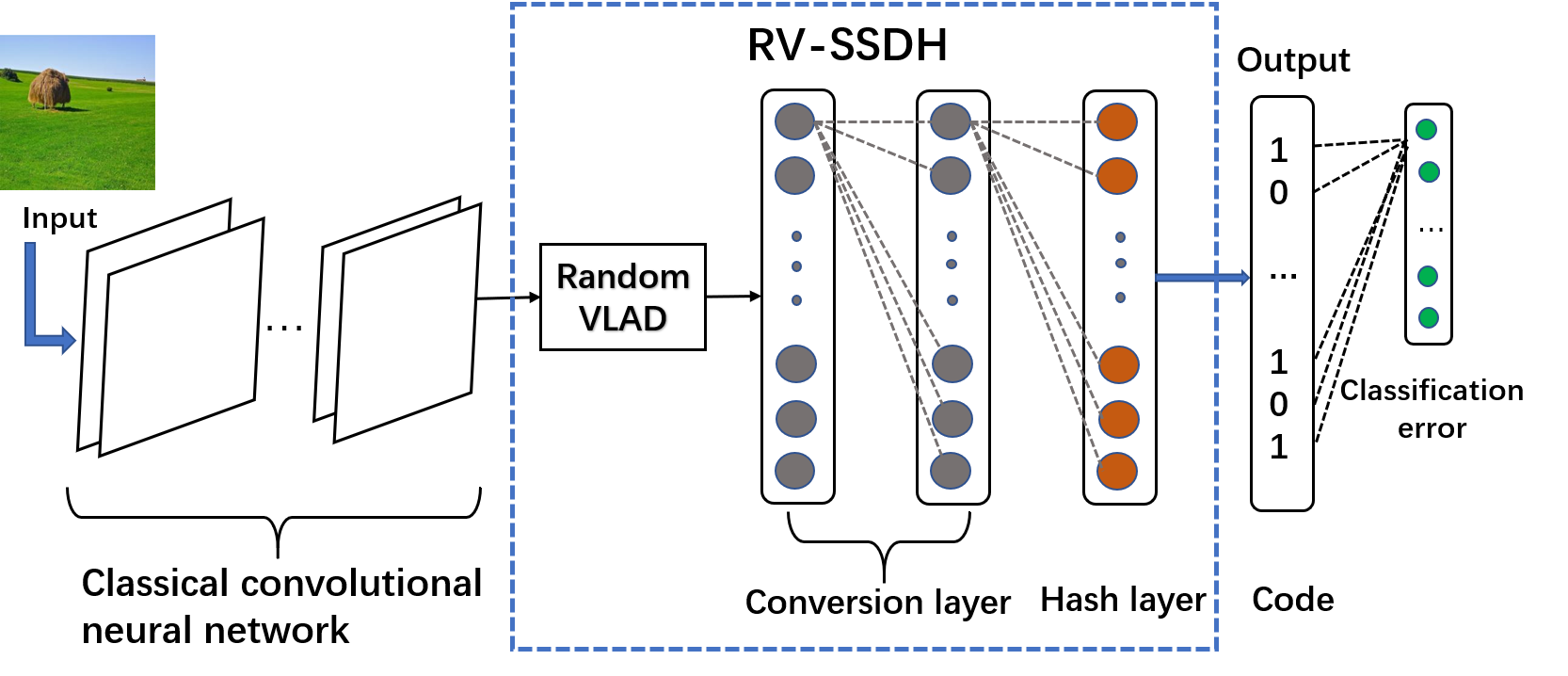}
\caption{Schematic diagram of the proposed RV-SSDH algorithm.}
\label{fig_rvssdh_diagram}
\vspace{-3mm}
\end{figure}

Given their advantages, aggregation and quantization can be combined. For example, BoF can be complemented by hashing~\cite{Jegou2008}; VLAD can be compressed by PQ~\cite{Jegou2010}. After the research community enters the deep learning era, both VLAD and hashing have developed their neural network versions. Two representative works are NetVLAD~\cite{Arandjelovic2018} and SSDH (supervised semantic-preserving deep hashing)~\cite{Yang2018}. While they outperform their counterparts based on hand-crafted features, a question has naturally arisen -- whether one can combine the advantages of the two. In this work, such an attempt is made. We propose a deep learning based hash algorithm which incorporates NetVLAD and SSDH. It is named RV-SSDH (random VLAD based SSDH). The core of the algorithm is a hash component that can be inserted into different neural networks (Fig.~\ref{fig_rvssdh_diagram}). The hash component consists of a random VLAD layer, a transform layer, and a hash layer. It is designed with the following properties: 1) RV-SSDH outperforms both SSDH and NetVLAD in terms of accuracy, speed, and compactness; 2) point-wise training is used for simplicity and versatility.
In particular, a random VLAD layer is used instead of the original NetVLAD. It works better in our scenario, and shows promising potentials as a general block in neural network design. To summarize, the contribution of this paper is two-fold:
\begin{itemize}
\item The proposed RV-SSDH is a state-of-the-art algorithm with O(N) complexity;
\item The proposed random VLAD block is an interesting alternative to the NetVLAD block. 
\end{itemize}
Extensive experiments have been performed. The results confirm that RV-SSDH outperforms SSDH and NetVLAD.

The rest of the paper is organized as follows. Section~\ref{sect_related_work} introduces related work in hashing. Section~\ref{sect_background} gives the background of VLAD. Section~\ref{sect_proposed_scheme} describes the proposed scheme. Section~\ref{sect_experiment} presents the experiment results. Section~\ref{sect_conclusion} concludes the work.

\section{Related work}
\label{sect_related_work}
Hashing, or robust Hashing, is also referred to as fingerprinting~\cite{Marc1996,Varna2011} or perceptual hashing~\cite{Weng2019}. Early techniques are data independent and use hand-crafted features. Many hash algorithms are approximately linear. They have a simple structure like the following:
\begin{align}
\mathbf{y}=\text{sgn}(\mathbf{W}^\intercal  \mathbf{x} + \mathbf{b}) \ ,
\label{eqn_hash_algorithm_linear}
\end{align}
where $\text{sgn}(\cdot)$ is the element-wise sign function, $\mathbf{x}$ is a feature vector, $\mathbf{W}$ is a transform matrix, and $\mathbf{b}$ is a bias vector. Existing approaches typically differ in the derivation of $\mathbf{W}$. Without learning, $\mathbf{W}$ can be generated randomly, but the performance is limited. A representative work is the locality-sensitive hashing (LSH)~\cite{Datar2004,Charikar2002,Slaney2008}. Learning based hash algorithms focus on computing $\mathbf{W}$ from training data, which can be divided into unsupervised and supervised categories. Unsupervised algorithms capture structure information in the feature data representation. They are versatile, but typically do not involve any ``semantic'' of data. Well-known methods include spectral hashing~\cite{Weiss2009}, iterative quantization~\cite{Gong2011}, k-means hashing~\cite{He2013}, and kernelized locality sensitive hashing~\cite{Kulis2009}.
In order to take semantics into account, supervised algorithms are developed. Typical examples are semi-supervised hashing~\cite{Wang2010}, kernel-based supervised hashing~\cite{Liu2012}, restricted Boltzmann machine based hashing~\cite{Torralba2008}, and supervised semantic-preserving deep hashing~\cite{Yang2018}. A comprehensive survey can be found in~\cite{Wang2016}.

In the deep learning era, the linear transformation in Eqn.~\eqref{eqn_hash_algorithm_linear} is replaced by neural networks with increased non-linearity and complexity~\cite{Zhang2015,Yang2018,Cakir2019}. The performance mainly depends on the structure of the network and the way of training. In particular, the use of ranking loss in image retrieval~\cite{Wang2014,Schroff2015,Arandjelovic2018} also brings improved performance in hashing~\cite{FangZhao2015,He2018}. On the other hand, new components, such as GAN~\cite{Cao2018} is beginning to be deployed.
Differing from previous work that relies on new training methods, this paper focuses on a new component in the main network structure called random VLAD. The proposed RV-SSDH is trained point-wise, thus takes relatively low computational cost.

\section{Background of VLAD}
\label{sect_background}
VLAD is related to the BoF~\cite{Sivic2003} representation. In the BoF representation, local features are detected from images, and described by descriptors such as SIFT~\cite{Lowe2004}. A local descriptor typically consists of a feature vector, a coordinate, a scale, an orientation, etc. Feature vectors usually go through vector quantization such as $k$-means~\cite{Lloyd1982}, and get encoded by a vocabulary of $K$ codewords, which are typically cluster centers in high-dimensional space. Let $\{\mathbf{f}_i\}^N_{i=1}$ denote feature vectors of an image. The BoF representation $\mathbf{d}$ is basically a histogram vector of codewords
\begin{align}
d_i=|\{\mathbf{f}_j\}|, \  \forall Q(\mathbf{f}_j)=\mathbf{c}_i, \\
i\in 1,\cdots,K \ , \ j\in 1,\cdots,N
\end{align}
where $Q(\cdot)$ is a vector quantization function, $d_i$ is the number of occurrences of the codeword $\mathbf{c_i}$, which is also called term frequency. 
The BoF generally shows satisfactory performance in image retrieval thanks to its robustness, but it has certain disadvantages: 1) 
it is typically a long sparse vector which needs compression; 2) too much information is lost by using a histogram, which decreases the discrimination power.
VLAD is an improvement on BoF. Let $\{\mathbf{v}_i\}^K_{i=1}$ denote the VLAD representation of an image. It tries to capture the neighbourhood of each codeword by the sum of associated descriptors
\begin{align}
\mathbf{v}_i=\sum_j \mathbf{f}_i, \  \forall Q(\mathbf{f}_j)=\mathbf{c}_i, \\
i\in 1,\cdots,K \ , \ j\in 1,\cdots,N
\end{align}
With increased discrimination, VLAD typically uses a smaller number of codewords. BoF and VLAD are still widely used in vision applications. VLAD also leads to extensions like NetVLAD~\cite{Arandjelovic2018} and multiscale NetVLAD~\cite{Shi2018}.

\section{The proposed scheme}
\label{sect_proposed_scheme}
The proposed RV-SSDH algorithm is a combination of conventional neural network layers and a novel hash component. A typical scheme is shown in Fig.~\ref{fig_rvssdh_diagram}, where convolutional layers are followed by an RV-SSDH layer and a classification layer. In order to facilitate transfer learning, convolutional layers of pre-trained networks can be used. For example, AlexNet~\cite{Krizhevsky2012} and VGG-F~\cite{Chatfield2014} are used in this work. In the following, we focus on the RV-SSDH component. It consists of three parts: a random VLAD layer, a transform layer, and a hash layer. They are explained in details below.
\begin{figure}[t]
\centering
\includegraphics[scale=0.3]{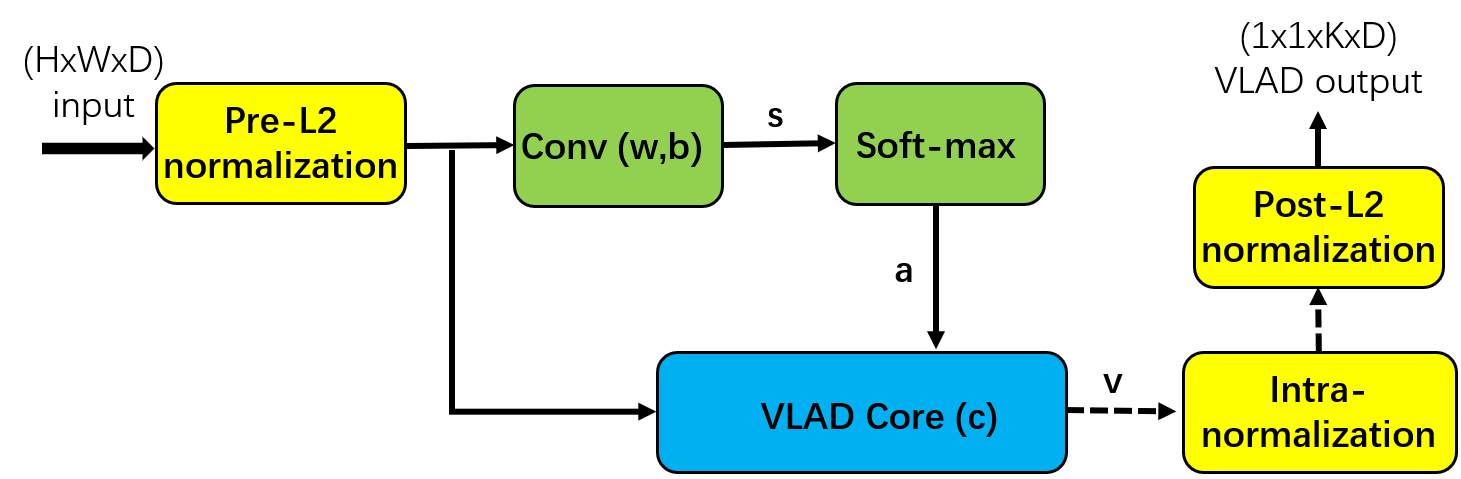}
\caption{The proposed random VLAD layer in comparison with NetVLAD. The yellow parts are used in NetVLAD.}
\label{fig_random_vlad}
\vspace{-3mm}
\end{figure}

\subsection{The random VLAD layer}

The random VLAD layer is a modified version of the one used in NetVLAD~\cite{Arandjelovic2018}, which is described here first. Figure~\ref{fig_random_vlad} gives a comparison between the two. In a NetVLAD network, the input to VLAD pooling is the $H \times W \times D$ output from a previous layer. It can be viewed as $H \times W$ $D$-dimensional ``local'' feature vectors.
The output size of VLAD core is $1 \times K \times D$. The $k$th output vector is defined as
\begin{equation}
\mathbf{y}_k=\sum^{N}_{i=1}{a_{ik}(\mathbf{x}_i-\mathbf{c}_k)} \ ,
\end{equation}
where $a_{k}(\mathbf{x}_i)=\mathbf{1}[Q(\mathbf{x}_i)=\mathbf{c}_k]$ indicates whether $\mathbf{x}_i$ is associated with the cluster center $\mathbf{c}_k$. In order to make it differentiable, $a_{k}(\cdot)$ is replaced with a soft assignment function
\begin{align}
a'_{k}(\mathbf{x}_i) & = \frac{\exp{(-\alpha||\mathbf{x}_i-\mathbf{c}_k||^2)}}{\sum^{K}_{j=1}{\exp{(-\alpha||\mathbf{x}_i-\mathbf{c}_j||^2)}}} \\
        & = \frac{\exp{(\mathbf{w}^\intercal_k \mathbf{x}_i+b_k)}}{\sum^{K}_{j=1}{\exp{(\mathbf{w}^\intercal_j \mathbf{x}_i+b_j)}}} \ ,
\end{align}
where $\mathbf{w}_k=2\alpha\mathbf{c}_k$ and $b_k=-\alpha||\mathbf{c}_k||^2$. In order to further improve flexibility, NetVLAD actually decouples $\{\mathbf{w}_k\}$, $\{b_k\}$ from $\{\mathbf{c}_k\}$ by setting them as three independent parameter sets.
The soft assignment is implemented by a convolution block followed by a softmax operation. The initial values of anchors $\{\mathbf{c}_k\}$ are obtained by applying $k$-means clustering to the input of VLAD core.

Besides the VLAD core, NetVLAD also contains a pre-L2 normalization layer, an intra normalization layer, and a post-L2 normalization layer. In practice, the last step can be skipped if cosine similarity is used for comparison.

The proposed scheme incorporates a random VLAD layer. It is similar to NetVLAD, but with the following differences:
\begin{itemize}
\item L2 and intra normalization are not used.
\item The anchors $\{\mathbf{c}_k\}$ are randomly initialized.
\end{itemize}
These modifications not only reduce algorithm complexity, but also improve retrieval performance, as shown by experiment results in Section~\ref{sect_netvlad_vs_random_vlad}.

\subsection{The transform layer}
The transform layer consists of two fully connected (FC) layers, each of which is followed by a rectified linear unit (ReLU) unit. 
The first FC layer converts the $K \times D$ output of VLAD core to a $D_1$-dimensional vector. The second FC layer further reduces the feature dimensionality from $D_1$ to $D_2$. In general, $D_2\leq D_1\leq K\times D$.
This part is motivated by the fact that some well-known networks typically have two FC layers after convolutional layers. They function as high-level feature extraction and transformation on top of low-level features. 

In our experiments, $D_1=D_2=1024$. According to the results in Sect.~\ref{sect_experiment_transform_layer}, the transform layer is not always necessary, but it generally helps to improve retrieval performance, especially for datasets with semantic gaps. We also find that, without preceding FC layers, the training of SSDH might not converge.

\subsection{The hash layer}
The hash layer originates from SSDH~\cite{Yang2018}. It consists of an FC layer and a sigmoid activation layer. The FC layer compresses the $D_2$-dimensional input to an $L$-dimensional feature vector. The feature vector is then binarized to derive an $L$-bit hash value. Since ideal binarization is not differentiable, the logistic sigmoid function is used as an approximation to facilitate back-propagation. This layer can be defined as
\begin{equation}
\hat{\mathbf{h}}=\sigma(\mathbf{W}^{\intercal}_h\mathbf{x}+\mathbf{b}_h) \label{eqn_hash_layer}
\end{equation}
where $\mathbf{W}_h$, $\mathbf{b}_h$ are the weight matrix and the bias vector respectively, and $\sigma(\cdot)$ is the sigmoid function $\sigma(z)=1/(1+\exp{(-z)})$ with the output range from $0$ to $1$.

In contrary to other hash algorithms, the hash layer is the last second layer (during training). It is assumed to contain latent attributes for classification. The elements of a hash value can be viewed as indicators of these attributes.

\subsection{The prediction layer}
The proposed hash algorithm is based on the architecture of SSDH. This architecture utilizes a classification problem to induce latent attributes, so a prediction layer is put after the hash layer.
The prediction layer is also an FC layer with sigmoid activation. It maps a hash value to class probabilities. The mapping is assumed to be a linear transformation, so the expression is similar to Eqn.~\eqref{eqn_hash_layer} but the bias term is ignored.
\begin{equation}
\hat{\mathbf{t}}=\sigma(\mathbf{W}^{\intercal}_c \hat{\mathbf{h}}) \ ,
\end{equation}
where $\hat{\mathbf{t}}$ is the predicted label vector.

\subsection{Optimization}
The proposed RV-SSDH is trained to generate similar hash values for similar (or relevant) images, in a similar way as SSDH. Let $N$ denote the number of samples. The basic objective function is based on classification error:
\begin{align}
\min:\{E_1=\sum_{i=1}^N L(\mathbf{t}_i,\hat{\mathbf{t}}_i)+\lambda \parallel \mathbf{W}_c \parallel^2\} \ ,
\end{align}
where $L(\cdot)$ represents the loss for the true label vector $\mathbf{t}_i$ and the predicted label vector $\hat{\mathbf{t}}_i$, and the constant $\lambda$ controls the relative weight of the regularization term.
In order to support different types of label information, the loss function takes the general form:
\begin{align}
L(\mathbf{t}_i,\hat{\mathbf{t}}_i)=-\sum_{j=1}^M l(t_{ij},\hat{t}_{ij}) \ ,
\end{align}
where $M$ is the number of classes, and $l(\cdot)$ depends on the application. This work mainly focuses on single-label classification, so the log loss is used
\begin{align}
l(t_{ij},\hat{t}_{ij})=t_{ij}\text{ln}(\hat{t}_{ij}) \ .
\end{align}
In order to make the hash output $\hat{\mathbf{h}}$ close to $0$ or $1$, another constraint is enforced:
\begin{align}
\max:\{E_2=\frac{1}{L}\sum_{i=1}^N \parallel \hat{\mathbf{h}}_i-\frac{1}{2}\mathbf{1} \parallel^p_{p}\}  \ ,
\end{align}
where $\hat{\mathbf{h}}$ is the continuous hash value, $\mathbf{1}$ is a vector of ``1''s, and $p \in\{1,2\}$.
Combining the constraints, the overall optimization problem is
\begin{align}
\text{arg}\min_\mathbb{W}: \{\alpha E_1-\beta E_2\} \ ,
\end{align}
where $\alpha, \beta$ are weight factors. Optimal parameters $\mathbb{W}$ of the network can be found by back propagation and stochastic gradient descent (SGD)~\cite{LeCun2015}. Since the algorithm is trained in a point-wise manner, the complexity of training is O(N).

In SSDH, there is another constraint that aims for equal probable bits:
\begin{align}
\min: \{E_3=\sum_{i=1}^N |\frac{1}{L}\hat{\mathbf{h}}^\intercal_i \mathbf{1} - 0.5|^p\} \ ,
\end{align}
This constraint is ignored in our implementation, for it has a minor impact on the retrieval performance~\cite{Yang2018}, and it potentially reduces the capacity of a hash space.

\subsection{Training and testing}
The training and testing of RV-SSDH is different. During training, network parameters are learned.
The prediction layer is only used in this phase, where the hash output $\hat{\mathbf{h}}$ is continuous.

During testing, hash values are first generated for a database and a query set. Then the retrieval performance is evaluated. In this phase, the prediction layer is removed, and $\hat{\mathbf{h}}$ is further quantized
\begin{align}
\mathbf{h}=(\text{sgn}(\hat{\mathbf{h}})+1)/2 \ ,
\end{align}
where $\mathbf{h}$ is a binary hash value. During retrieval, the Hamming distance is used for comparing hash values.

\section{Experiment results}
\label{sect_experiment}
Comprehensive experiments are performed to evaluate RV-SSDH. First, retrieval performance is examined together with classification performance. Then algorithm complexity is measured in terms of retrieval speed and training speed. Initial results are obtained for small datasets and compared with baselines. Following a large-scale retrieval test, more experiments are carried out to study the effects of random VLAD. Finally, RV-SSDH is compared with some more algorithms from state-of-the-art.
The detailed results are described below. Figures are best viewed in color.

\subsection{Datasets and evaluation metrics}
Three datasets are used in the paper: MNIST~\cite{LeCun1998}, CIFAR-10~\cite{Krizhevsky2009}, and Places365~\cite{Zhou2018}. MNIST is a gray image dataset of handwritten digits (from 0 to 9). CIFAR-10 is a dataset of color images in these classes: airplane, automobile, dog, cat, bird, deer, frog, horse, ship, truck. They both contain ten classes and 6000 images per class. The image sizes are $28\times 28$ and $32 \times 32$ respectively. For these two datasets, we use 10000 images for validation and the rest for training. Places365 is a large-scale dataset of 365 scene categories. We use its training set which contains more than 1.8 million images: 80\% is used for training and the rest for validation. In order to reveal the performance gain brought by RV-SSDH, no data augmentation is used.

The three datasets have different purposes: MNIST is a relatively simple dataset without much semantics; CIFAR-10 is more difficult for it has severe semantic gaps; Places365 is the most challenging.

Algorithms are tested in a retrieval scenario. For MNIST and CIFAR-10, the validation set is used as a database, and 1000 items from the database are randomly selected as queries; for Places365, 30000 images from the validation set are used as a database, and 3000 images from the database are randomly selected as queries.
The retrieval performance is measured by two metrics: the precision-recall (P-R) curve and the mean average precision (mAP). For a query, precision and recall are defined as
\begin{align}
\text{precision}=\frac{\text{number of retrieved relevant items}}{\text{number of retrieved items}} \ , \\
\text{recall}=\frac{\text{number of retrieved relevant items}}{\text{number of total relevant items}} \ .
\end{align}
Different trade-offs can be achieved by adjusting the number of retrieved items. A P-R curve is plotted by averaging over all queries. The mAP is defined as area under the curve which represents overall retrieval performance.

\subsection{The baselines}
The SSDH~\cite{Yang2018} and the NetVLAD~\cite{Arandjelovic2018} algorithms are the main baselines in this work. Since RV-SSDH is a pluggable component, the actual implementation of a complete network also depends on the preceding layers. For a particular setting of convolutional and FC layers, denoted by CNN, the following baselines are considered
\begin{itemize}
\item CNN;
\item CNN+(FC)+SSDH;
\item CNN+NetVLAD;
\item CNN+RV-SSDH.
\end{itemize}
Specifically, three CNNs are used: the first two are the well-known AlexNet~\cite{Krizhevsky2012} and VGG-F~\cite{Chatfield2014}, which can test RV-SSDH in a transfer learning scenario; the third one is a small custom network defined in Table~\ref{tab_toynet}, which we call ToyNet. When AlexNet or VGG-F is used, parameters of pre-trained models (based on ImageNet) are loaded into the convolutional layers, while ToyNet is trained from scratch. Note that when a CNN is used alone, it includes convolutional layers and possibly two FC layers (for AlexNet/VGG-F); when a CNN is used together with another block, only its convolutional layers are used. For example, AlexNet+RV-SSDH means the convolutional layers of AlexNet (conv1--conv5) are combined with RV-SSDH. A special case is AlexNet/VGG-F+FC+SSDH, where two FC layers are added in the middle. This is because we find that without the FC layers it is difficult to make the training converge.


\begin{table}[t]
\centering
\caption{ToyNet: a small custom convolutional neural network.}
\includegraphics[scale=0.45]{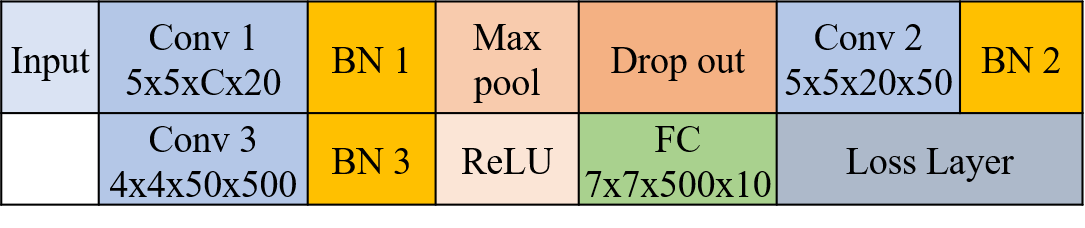}
\label{tab_toynet}
\vspace{-3mm}
\end{table}

It should also be noted that the original NetVLAD uses triplet-based training~\cite{Arandjelovic2018,Schroff2015}. It is modified to use point-wise training in our implementation. This modification not only guarantees fair comparison with others, but also reveals the performance of NetVLAD in a more general setting.

For the same test, the same amount of epochs (typically 50) is used for all candidate algorithms.
For SSDH and RV-SSDH, hash lengths from 8 to 128 bits are mainly considered.

\subsection{Retrieval performance}
\begin{figure}[t]
\centering
\includegraphics[scale=0.5]{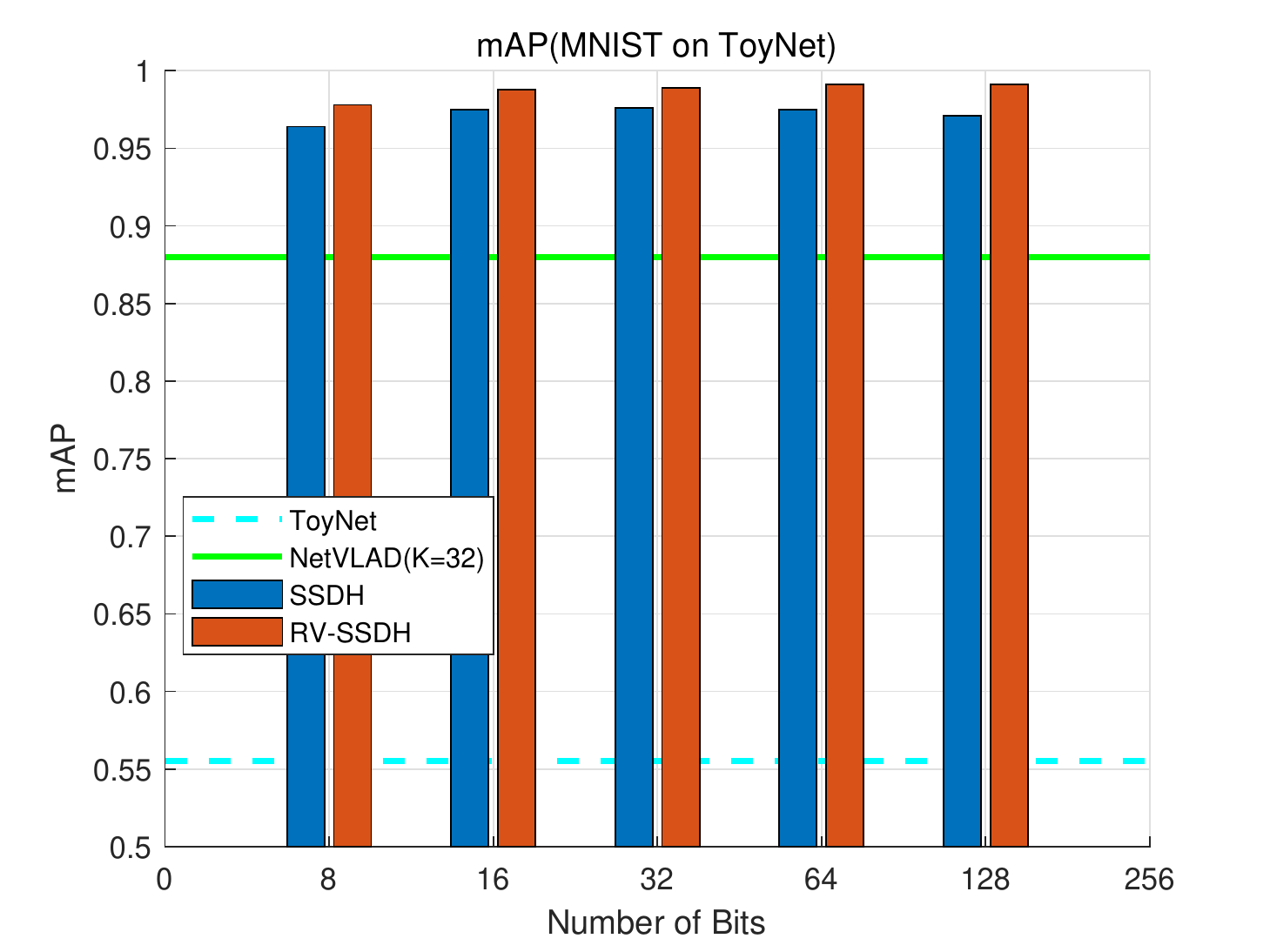}
\caption{mAP comparison (MNIST, ToyNet).}
\label{fig_map_mnist_toynet}
\vspace{-3mm}
\end{figure}

Figure~\ref{fig_map_mnist_toynet} shows a comparison of mAP values, where the dataset is MNIST and the base network is ToyNet. For SSDH and RV-SSDH, the mAP varies with the hash length; for NetVLAD, only the best mAP is shown; for ToyNet, the mAP is constant. One can see that using ToyNet alone leads to an mAP around $0.55$, while NetVLAD can boost it to $0.88$. The large difference confirm the effectiveness of NetVLAD. On the other hand, SSDH performs even better, and RV-SSDH gives the highest mAP. The observation gives a basic ranking: RV-SSDH$>$SSDH$>$NetVLAD$>$ToyNet.

\begin{figure}
\centering
\includegraphics[scale=0.5]{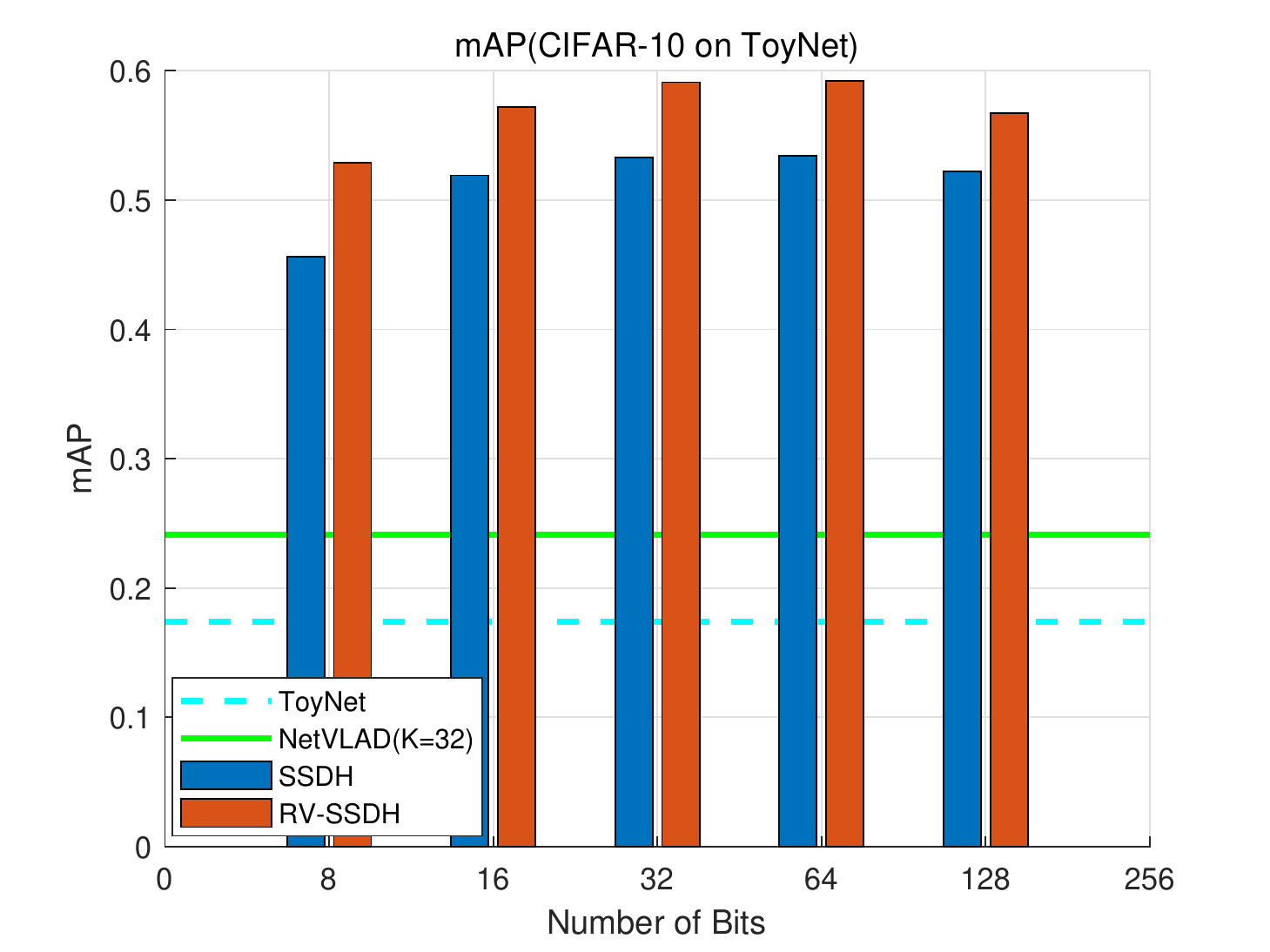}
\caption{mAP comparison (CIFAR-10, ToyNet).}
\label{fig_map_cifar10_toynet}
\vspace{-3mm}
\end{figure}

Figure~\ref{fig_map_cifar10_toynet} shows a comparison of mAP values, where the dataset is CIFAR-10 and the base network is ToyNet. A similar trend is observed, but the mAP gain over ToyNet is not as large as in Fig.~\ref{fig_map_mnist_toynet}, especially for NetVLAD. This is because CIFAR-10 is more difficult than MNIST and ToyNet is not a sophisticated network. The advantage of RV-SSDH becomes more noticeable in this case. 

\begin{figure}
\centering
\includegraphics[scale=0.5]{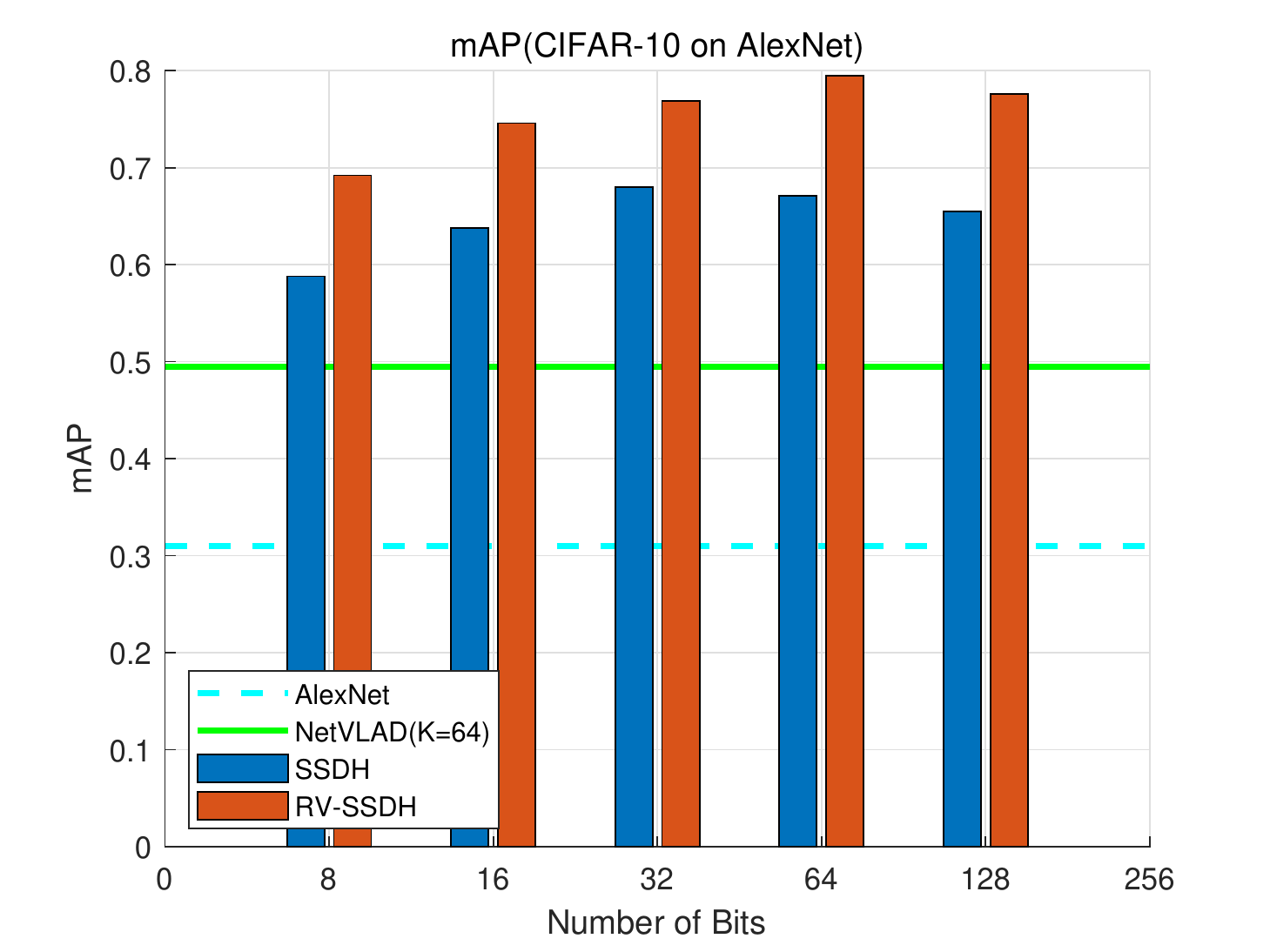}
\caption{mAP comparison (CIFAR-10, AlexNet).}
\label{fig_map_cifar10_alexnet}
\vspace{-3mm}
\end{figure}

Figure~\ref{fig_map_cifar10_alexnet} shows a comparison of mAP values, where the dataset is CIFAR-10 and the base network is AlexNet. The trend stays the same. Since a more sophisticated base network is used, the mAP is generally much improved, compared with ToyNet. RV-SSDH still performs the best, with an approximate margin of $0.1$ above SSDH.

The results from Fig.~\ref{fig_map_mnist_toynet} to Fig.~\ref{fig_map_cifar10_alexnet} are consistent, so one can basically conclude that RV-SSDH has the best retrieval performance among the candidate algorithms. 
These figures also provide some other insights. For example, the mAP typically increases with the hash length $L$ (especially when $L$ is small), but it might decrease when $L$ is too large. This could be a consequence of insufficient training -- a larger $L$ requires more network parameters while the number of epochs is fixed, so a large $L$ does not necessarily guarantee better discrimination power. The results show that $L=64$ typically works best.
Compared with others, NetVLAD seems more sensitive to the choice of base network and the dataset.

\begin{figure}
\centering
\includegraphics[scale=0.5]{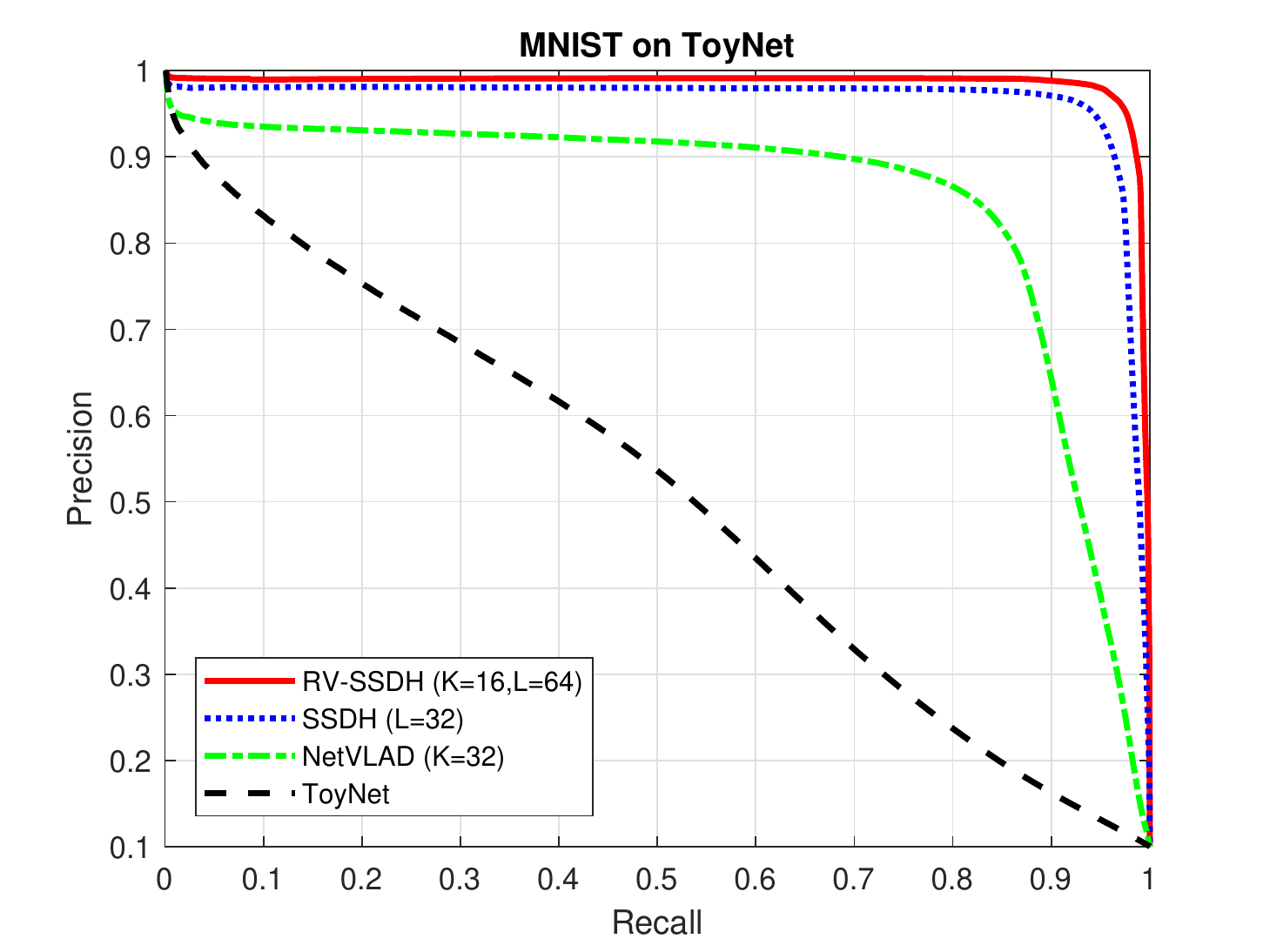}
\caption{P-R curve comparison (MNIST, ToyNet).}
\label{fig_pr_mnist_toynet}
\vspace{-3mm}
\end{figure}

\begin{figure}
\centering
\includegraphics[scale=0.5]{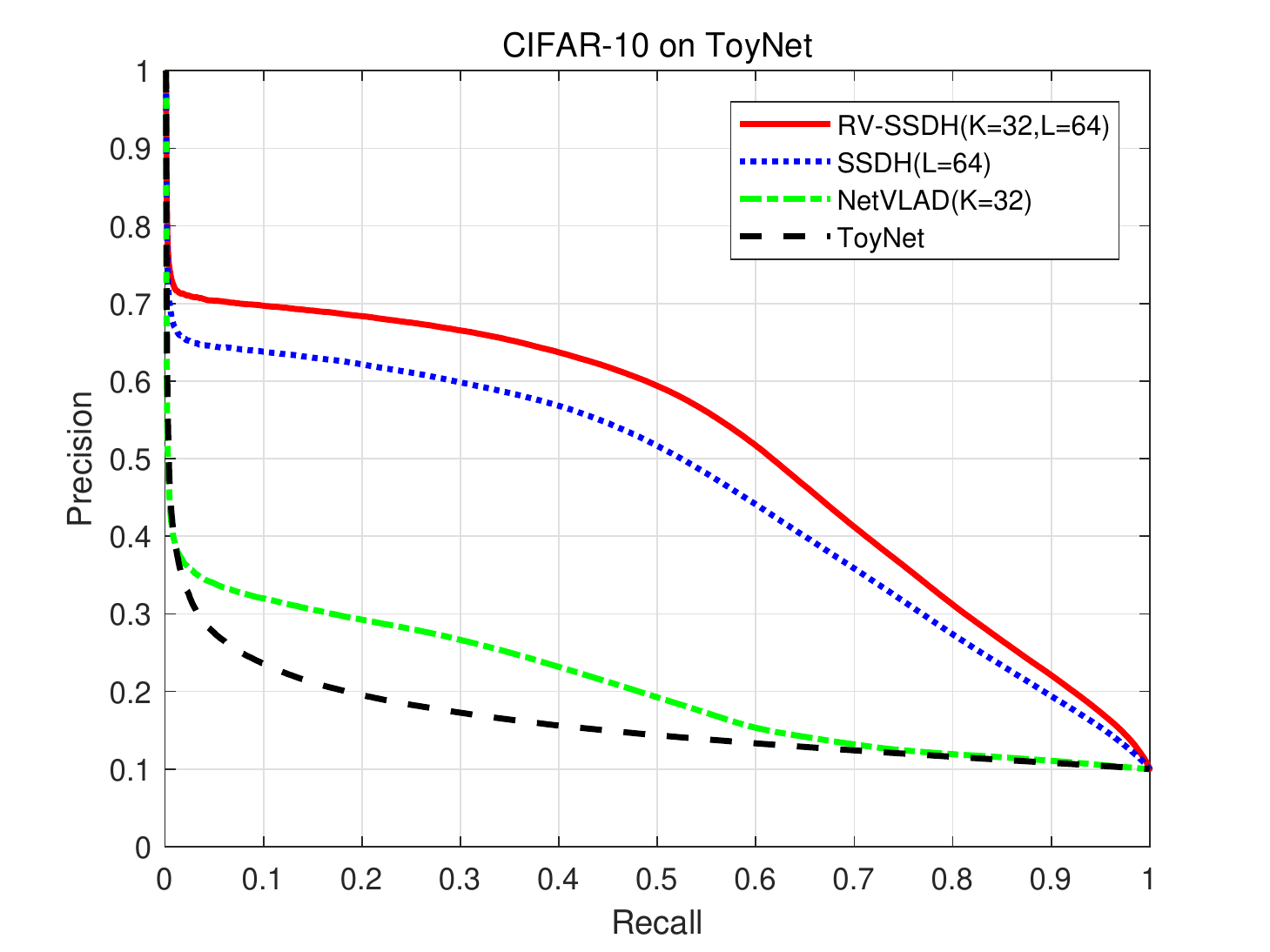}
\caption{P-R curve comparison (CIFAR-10, ToyNet).}
\label{fig_pr_cifar10_toynet}
\vspace{-3mm}
\end{figure}

\begin{figure}[t]
\centering
\includegraphics[scale=0.5]{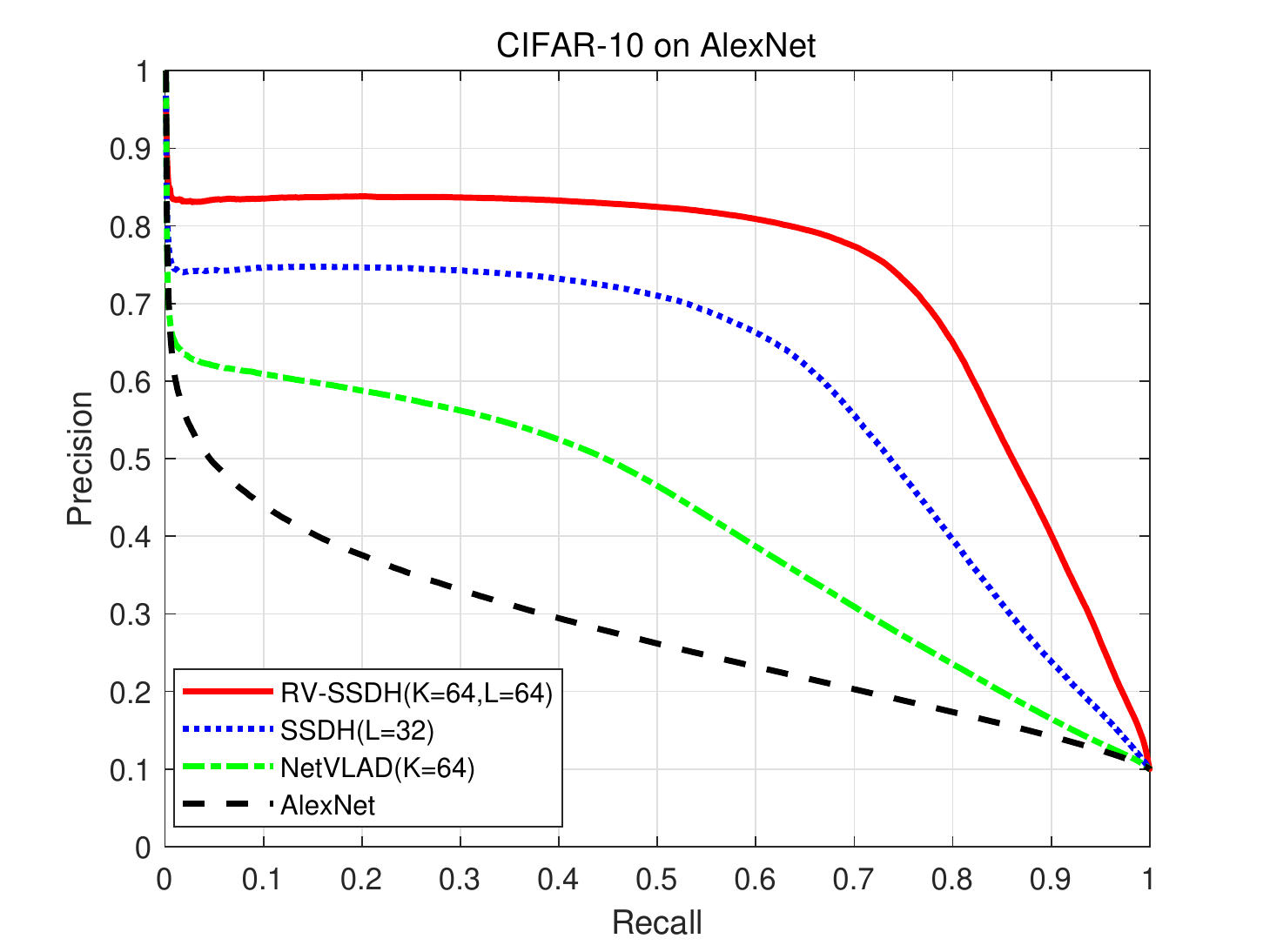}
\caption{P-R curve comparison (CIFAR-10, AlexNet).}
\label{fig_pr_cifar10_alexnet}
\vspace{-3mm}
\end{figure}

The mAP represents overall retrieval performance. More details and trade-offs can be found in the precision-recall curve. Figures~\ref{fig_pr_mnist_toynet}--\ref{fig_pr_cifar10_alexnet} show some comparisons of P-R curves for MNIST and CIFAR-10. In these figures, the parameter(s) of each curve corresponds to the algorithm's highest mAP. For example, in Fig.~\ref{fig_pr_mnist_toynet}, for SSDH and MNIST, the best mAP is achieved by $L=32$, which is outperformed by RV-SSDH with $K=16$ and $L=64$; for NetVLAD, the best mAP is achieved when $K=32$, whose curve is significantly above ToyNet's but clearly below SSDH's.

Note that in Fig.~\ref{fig_pr_cifar10_toynet}, the P-R curve of NetVLAD is not always above ToyNet's. This explains why NetVLAD only has a small advantage over ToyNet in Fig.~\ref{fig_map_cifar10_toynet}. These results again confirm that RV-SSDH has superior retrieval performance.

\subsection{Classification performance}
Since the proposed RV-SSDH is trained in a classification framework (recall that there is a prediction layer after the hash layer), a question is whether a hash value is also suitable for classification. 

To answer this question, Table~\ref{tab_top1_mnist_toynet} shows a comparison on the Top-1 error rate during validation, where the dataset is MNIST and the base network is ToyNet. In the table, the smallest and the two largest rates are marked in bold. It is interesting to see that NetVLAD gives the largest value (SSDH gives the second largest), and the general performance ranking is: RV-SSDH$>$SSDH$>$ToyNet$>$NetVLAD, which is different from the retrieval cases. Compared with Fig.~\ref{fig_map_mnist_toynet}--\ref{fig_pr_cifar10_alexnet}, the results show a difference between classification and retrieval. In other words, good retrieval performance does not guarantee good classification performance, or vice versa. Nevertheless, for RV-SSDH, the behaviour is consistent.

More results are shown in Table~\ref{tab_top1_cifar10_toynet}--\ref{tab_top1_cifar10_alexnet}, and similar patterns can be observed. Note that in Table~\ref{tab_top1_cifar10_alexnet} AlexNet performs better than SSDH but worse than RV-SSDH. The best accuracy achieved by RV-SSDH on CIFAR-10 with AlexNet is 86.19\%, which is close to the accuracy (89\%) in the AlexNet paper~\cite{Krizhevsky2012}. Considering that no data augmentation is used in our experiments and our AlexNet's accuracy is 82.61\%, we conclude that RV-SSDH is likely to give a significant boost in classification performance.

Note that the error rates of SSDH and RV-SSDH are computed using unbinarized hash values (i.e. $\hat{\mathbf{h}}$). Therefore the above results only prove that the continuous RV-SSDH is useful for classification. Although the quantization error is generally small, the actual performance of binary hash values in classification is left for future investigation.

\begin{table}
\centering
\caption{Top-1 error comparison (MNIST, ToyNet).}
\begin{tabular}{l|lllll}
\hline
\multirow{2}{*}{Method}	& \multicolumn{5}{c}{error rate vs. hash length}	\\
	\cline{2-6}		
	& 8 bits&	16 bits	& 32 bits & 64 bits	& 128 bits \\
\hline
ToyNet	& \multicolumn{5}{c}{0.0126}	 \\			
NetVLAD	& \multicolumn{5}{c}{\bf{0.0474}} \\		
SSDH	&\bf{0.0147}	&0.0129	&0.0117	&0.0118	&0.0116 \\
RV-SSDH	&0.0101	&0.0089	&\bf{0.0087}	&0.0093	&0.0088 \\
\hline
\end{tabular}
\label{tab_top1_mnist_toynet}
\vspace{-3mm}
\end{table}


\begin{table}[t]
\centering
\caption{Top-1 error comparison (CIFAR-10, ToyNet).}
\begin{tabular}{l|lllll}
\hline
\multirow{2}{*}{Method} & \multicolumn{5}{c}{error rate vs. hash length}	\\
\cline{2-6}				
       &	8 bits	& 16 bits	& 32 bits	& 64 bits	& 128 bits\\
\hline
ToyNet	& \multicolumn{5}{c}{0.2802} \\			
NetVLAD	& \multicolumn{5}{c}{\bf{0.5787}}\\			
SSDH	&\bf{0.3127} & 0.2757 & 0.2675 & 0.2724 & 0.2714\\
RV-SSDH	&0.2780 & 0.2453 & 0.2375 & \bf{0.2313} & 0.2357\\
\hline
\end{tabular}
\label{tab_top1_cifar10_toynet}
\vspace{-3mm}
\end{table}


\begin{table}[t]
\caption{Top-1 error comparison (CIFAR-10, AlexNet).}
\begin{tabular}{l|lllll}
\hline
\multirow{2}{*}{Method}	&\multicolumn{5}{c}{error rate vs. hash length}\\
\cline{2-6}
	&8 bits	& 16 bits	&32 bits	&64 bits	&128 bits\\
\hline
Alexnet	&\multicolumn{5}{c}{0.1739} \\				
NetVLAD	&\multicolumn{5}{c}{\bf{0.2846}}\\	
SSDH	&\bf{0.2306} & 0.2158 & 0.2029 & 0.2049 & 0.2079\\
RV-SSDH	&0.1637 & 0.1413 & 0.1472 & \bf{0.1381} & 0.1407\\
\hline
\end{tabular}
\label{tab_top1_cifar10_alexnet}
\vspace{-3mm}
\end{table}

\subsection{Complexity comparison}
\begin{figure}
\centering
\includegraphics[scale=0.5]{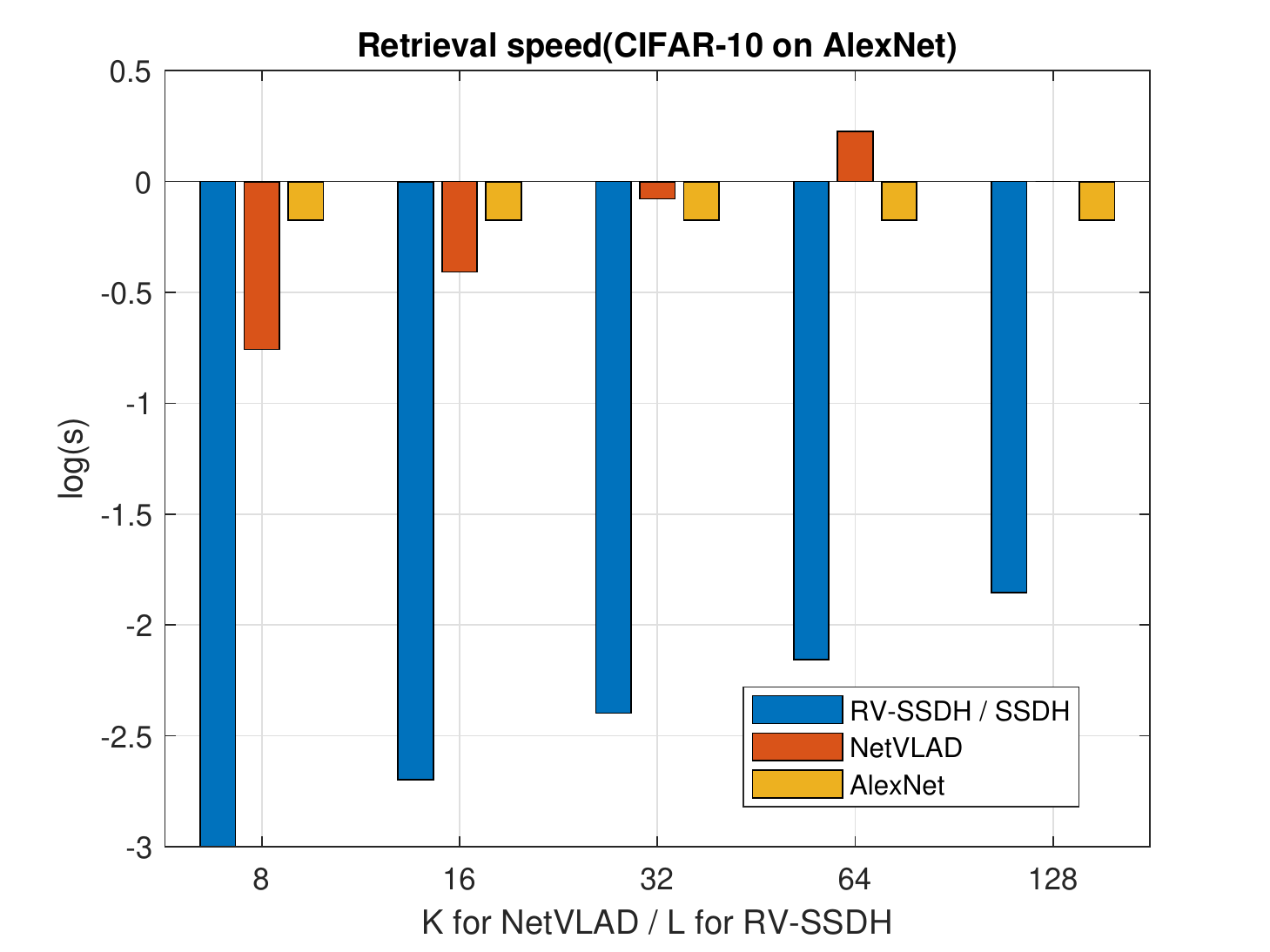}
\caption{Retrieval speed comparison (seconds/query) in log scale (CIFAR-10, AlexNet).}
\label{fig_speed_cifar10_alexnet}
\vspace{-3mm}
\end{figure}

The complexity of RV-SSDH can be evaluated in two aspects: 1) the retrieval speed; 2) the training speed. Some quantitative results are given in this section.  The experiments are performed on a computer with Intel i7-8700K CPU, 16G memory, and Nvidia GTX1080 GPU. 
Figure~\ref{fig_speed_cifar10_alexnet} shows a comparison of retrieval speed (seconds per query) in log scale, where the dataset is CIFAR-10 and the base network is AlexNet. For RV-SSDH and SSDH, the retrieval speed is the same and the hash length $L$ is varied; for NetVLAD, the centroid number $K$ is varied, and the output length is $K\times D=256K$; for AlexNet, the classification layer is removed, and the output length is $4096$.

\begin{figure}
\centering
\includegraphics[scale=0.5]{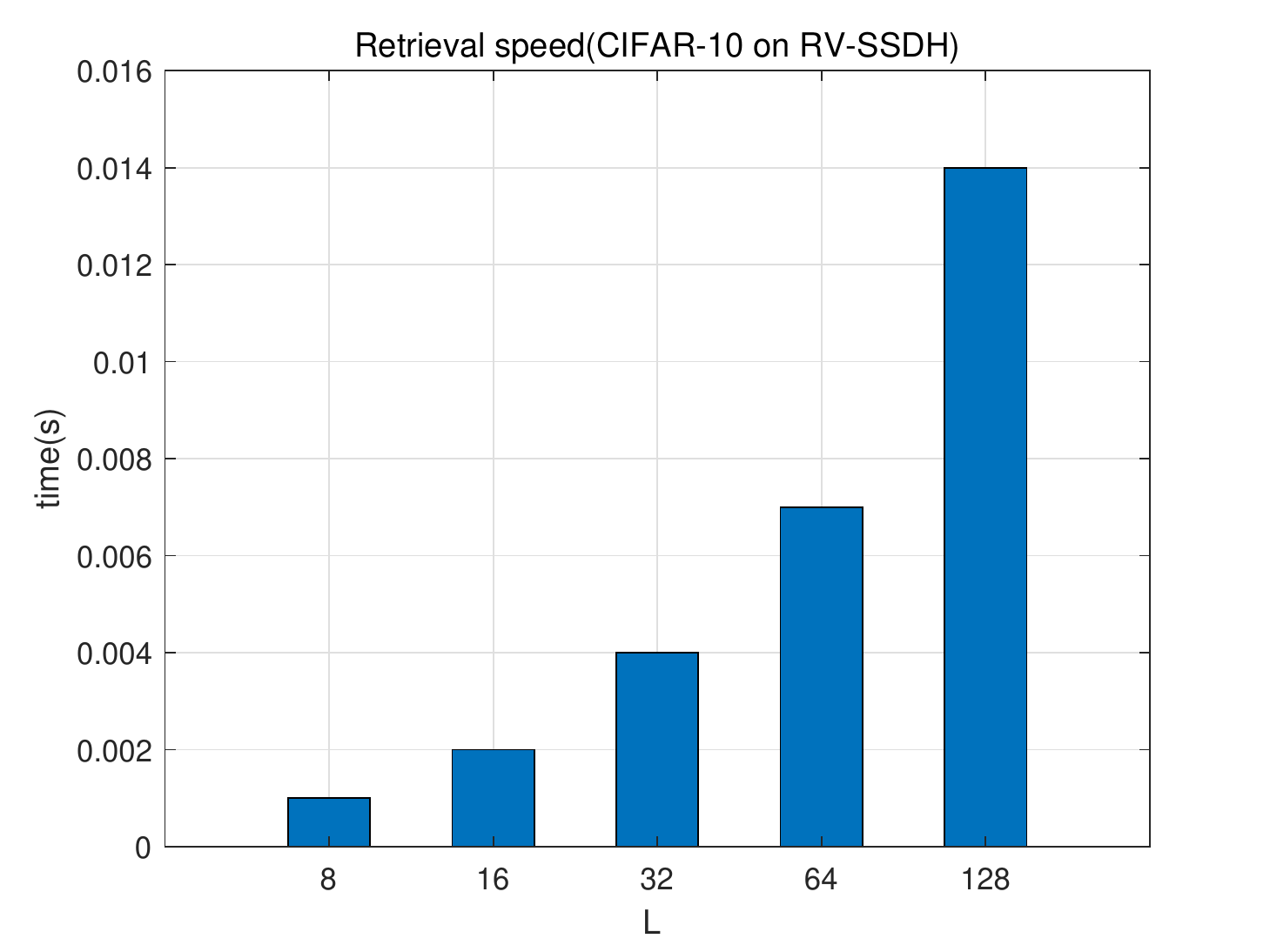}
\caption{Retrieval speed (seconds/query) of RV-SSDH (CIFAR-10, AlexNet).}
\label{fig_speed_rvssdh_cifar10_alexnet}
\vspace{-3mm}
\end{figure}

Note that the distance metric for retrieval is different for different algorithms: RV-SSDH/SSDH uses Hamming distance; NetVLAD uses Cosine distance; AlexNet uses Euclidean distance. This is why NetVLAD is faster than AlexNet for the same output size. In the NetVLAD paper~\cite{Arandjelovic2018}, the authors also propose to compress the output with PCA~\cite{Bishop2006} to $4096$ dimensions. That corresponds to the $K=16$ case. To conclude, RV-SSDH and SSDH are always the fastest; NetVLAD is faster than AlexNet for small $K$. Figure~\ref{fig_speed_rvssdh_cifar10_alexnet} shows the retrieval speed of RV-SSDH in linear scale. The actual speed is approximately linear in the hash length.

Table~\ref{tab_speed_training} shows a comparison of training speed, which is represented by the number of processed images per second (Hz). It is obvious that AlexNet is the fastest, because fewest layers are used. The speed of SSDH is about half of AlexNet. The hash length $L$ is not shown in the table because we find that the training speed is insensitive to $L$. On the other hand, for RV-SSDH and NetVLAD, the speed actually depends on the anchor number $K$. For a small $K$, RV-SSDH can be faster than SSDH and get close to AlexNet. This is because the VLAD core reduces the complexity of subsequent FC layers. In general, the speed of NetVLAD is about half of RV-SSDH. The speed gain of RV-SSDH comes from removing the normalization layers. To conclude, RV-SSDH is almost always a good choice in terms of retrieval accuracy and complexity.

\begin{table}
\centering
\caption{Training speed comparison (CIFAR-10, AlexNet).}
\begin{tabular}{c|llll}
\hline
Method & AlexNet	& NetVLAD &	SSDH & RV-SSDH\\
\hline
 & \multirow{4}{*}{987} &	410 (K=8)	& \multirow{4}{*}{540} & 900 (K=8) \\
Speed &   &               360 (K=16)    &    &                   800 (K=16)\\
(Hz)    & &                           310 (K=32)   &    &                    620 (K=32)\\
       & &                        240 (K=64)  &      &                   415 (K=64)\\
\hline
\end{tabular}
\label{tab_speed_training}
\vspace{-3mm}
\end{table}

\subsection{Large-scale retrieval}
Besides MNIST and CIFAR-10, a much larger dataset Places365~\cite{Zhou2018} is also used to evaluate RV-SSDH. Since previous results show that NetVLAD generally performs worse than SSDH, it is no longer considered in this test. Table~\ref{tab_map_places365} shows a mAP comparison between RV-SSDH and SSDH for hash lengths 256 and 512 (larger hash lengths are used here for the dataset is much larger than before). Two base networks are used: AlexNet and VGG-F~\cite{Chatfield2014}. Although the margin is not as large as in the CIFAR-10 case, RV-SSDH still outperforms SSDH by 2\%--4.9\% in mAP.
We also note that VGG-F works better than AlexNet, and 256-bit length works better than 512-bit length (perhaps due to insufficient training).

\begin{table}[t]
\centering
\caption{mAP comparison with Places365.}
\begin{tabular}{l|ll}
\hline
\multirow{2}{*}{method} & \multicolumn{2}{c}{mAP} \\
\cline{2-3}
& VGG-F & AlexNet \\
\hline
RV-SSDH (K=32,L=256)	&\bf{0.2023}  & \bf{0.1904} \\
RV-SSDH (K=32,L=512)	&0.1835	& 0.1679 \\
SSDH (L=256)		& 0.1822	& 0.1413 \\
SSDH (L=512)		& 0.1595	& 0.1292 \\
\hline
\end{tabular}
\label{tab_map_places365}
\vspace{-3mm}
\end{table}

%
%

\subsection{The choice of VLAD parameters}
\begin{figure}
\centering
\includegraphics[scale=0.5]{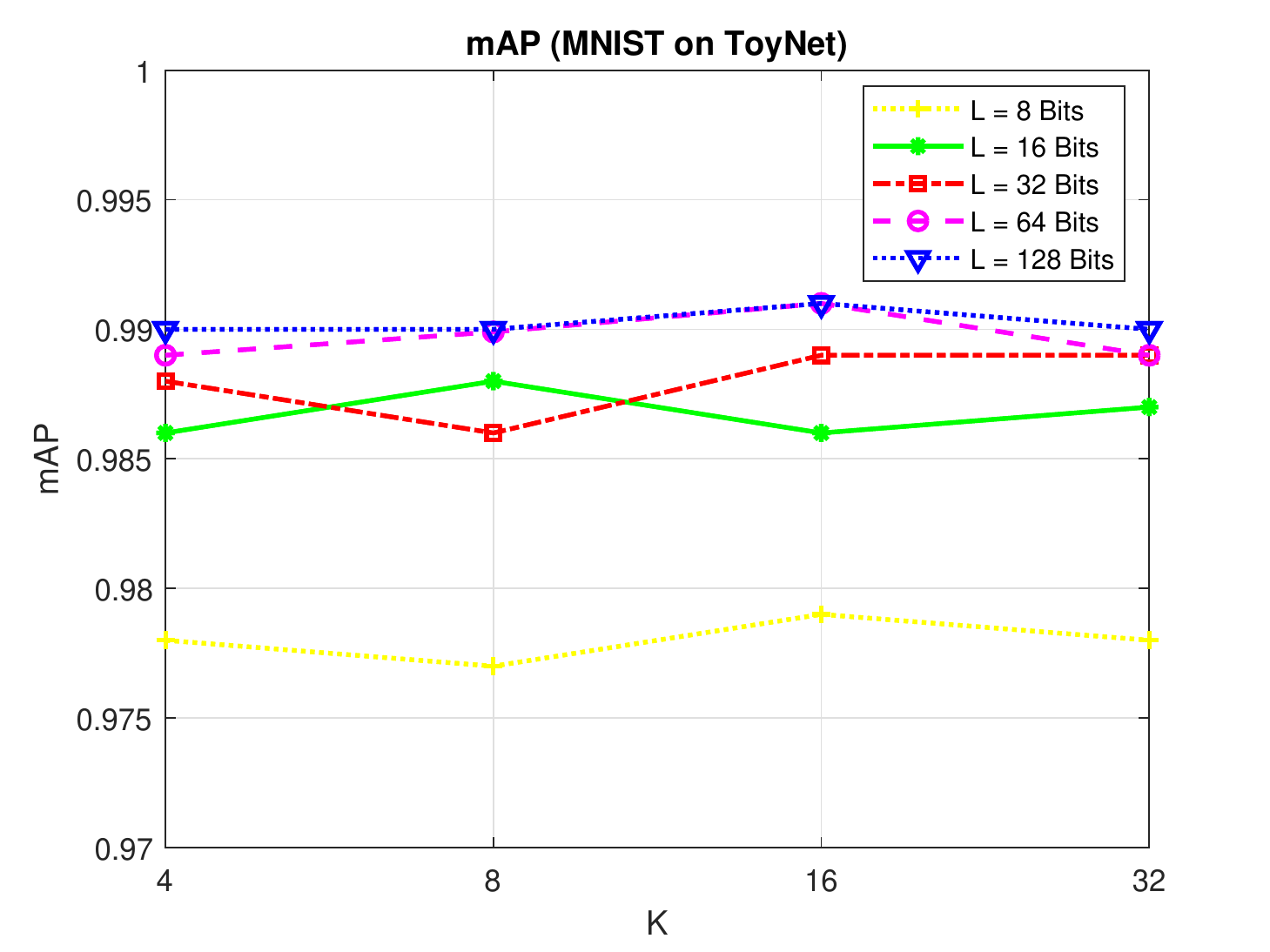}
\caption{mAP comparison for RV-SSDH (MNIST, ToyNet).}
\label{fig_map_mnist_toynet_rvssdh}
\end{figure}

\begin{figure}
\centering
\includegraphics[scale=0.5]{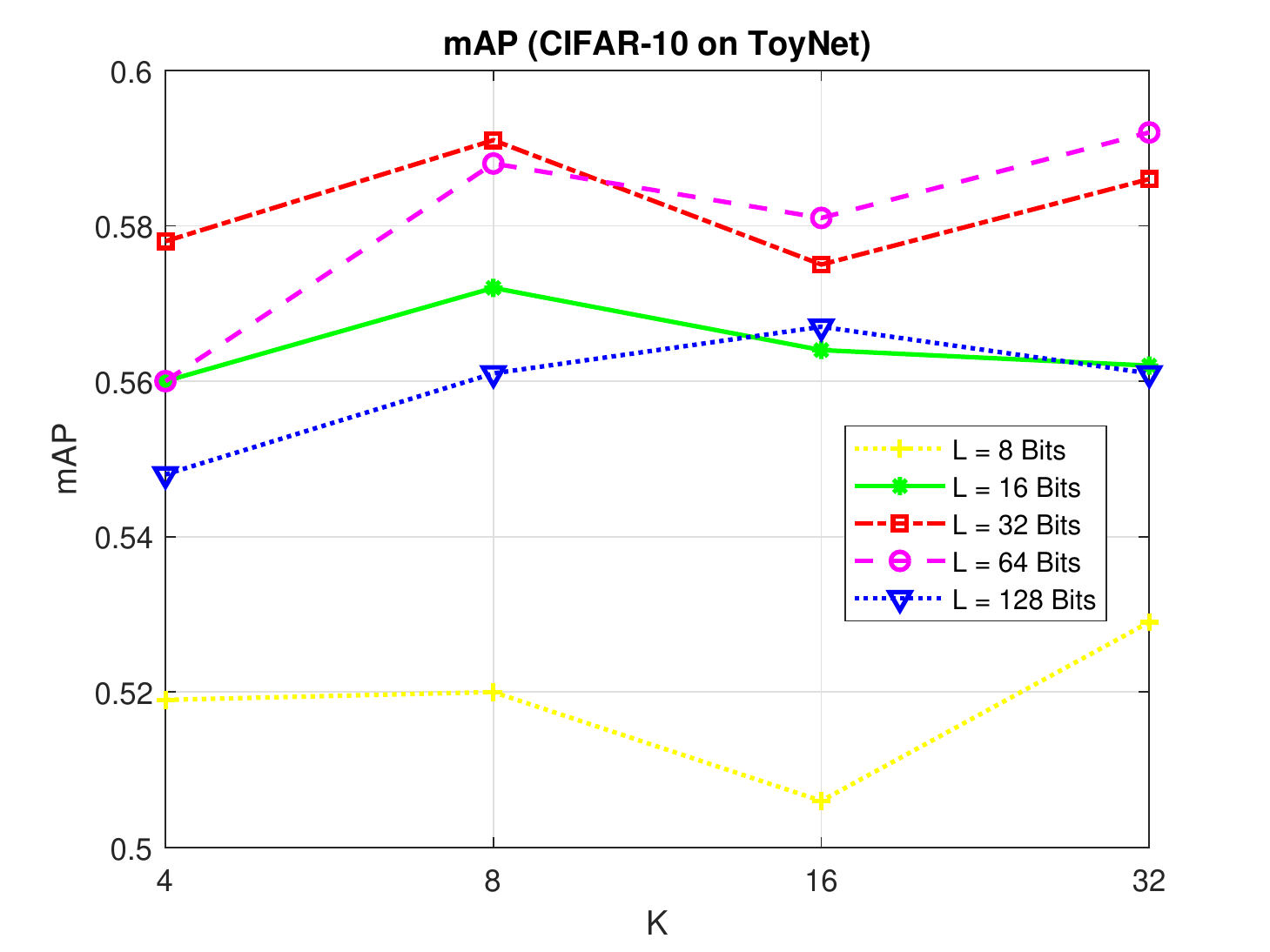}
\caption{mAP comparison for RV-SSDH (CIFAR-10, ToyNet).}
\label{fig_map_cifar10_toynet_rvssdh}
\end{figure}

\begin{figure}
\centering
\includegraphics[scale=0.5]{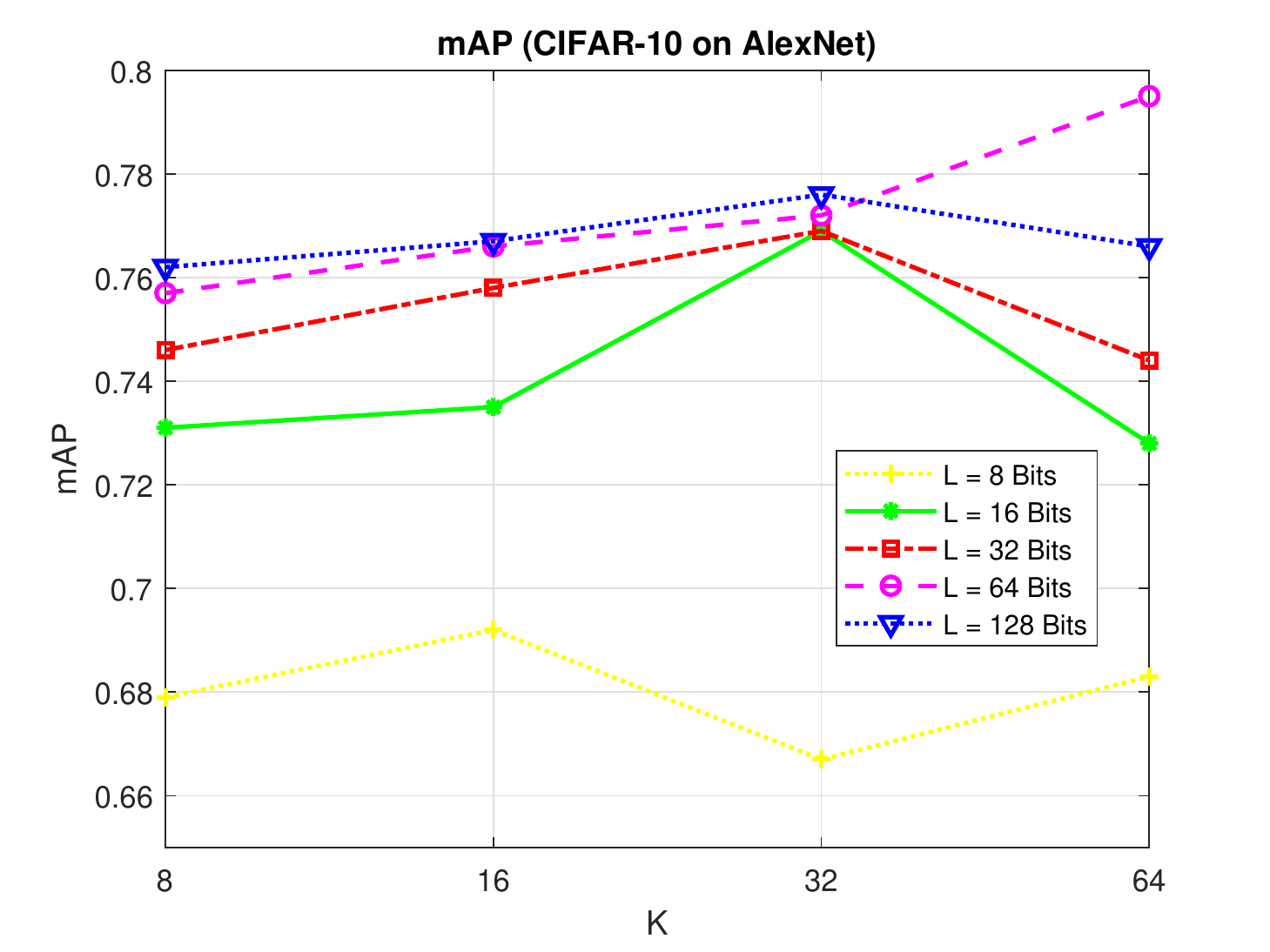}
\caption{mAP comparison for RV-SSDH (CIFAR-10, AlexNet).}
\label{fig_map_cifar10_alexnet_rvssdh}
\end{figure}

RV-SSDH has parameters $L$ and $K$. The choice of $L$ typically depends on the number of classes, while $K$ controls the level of aggregation. By varying $K$, $L$ and comparing the mAP, some tests are performed to find suitable values for these parameters. While the hash length $L$ typically depends on the dataset size and the number of classes, it is not straightforward to see the best value of $K$. According to the results in Figures~\ref{fig_map_mnist_toynet_rvssdh}--\ref{fig_map_cifar10_alexnet_rvssdh}, one can see that $K=16, 32$ and $L=64$ are generally good choices for the tested datasets. Since $K$ also controls the complexity of the algorithm (see Table~\ref{tab_speed_training}), our rule of thumb is that $K$ should not be too large.

\subsection{The effects of random VLAD}
\label{sect_netvlad_vs_random_vlad}
\begin{figure}[t]
\centering
\includegraphics[scale=0.5]{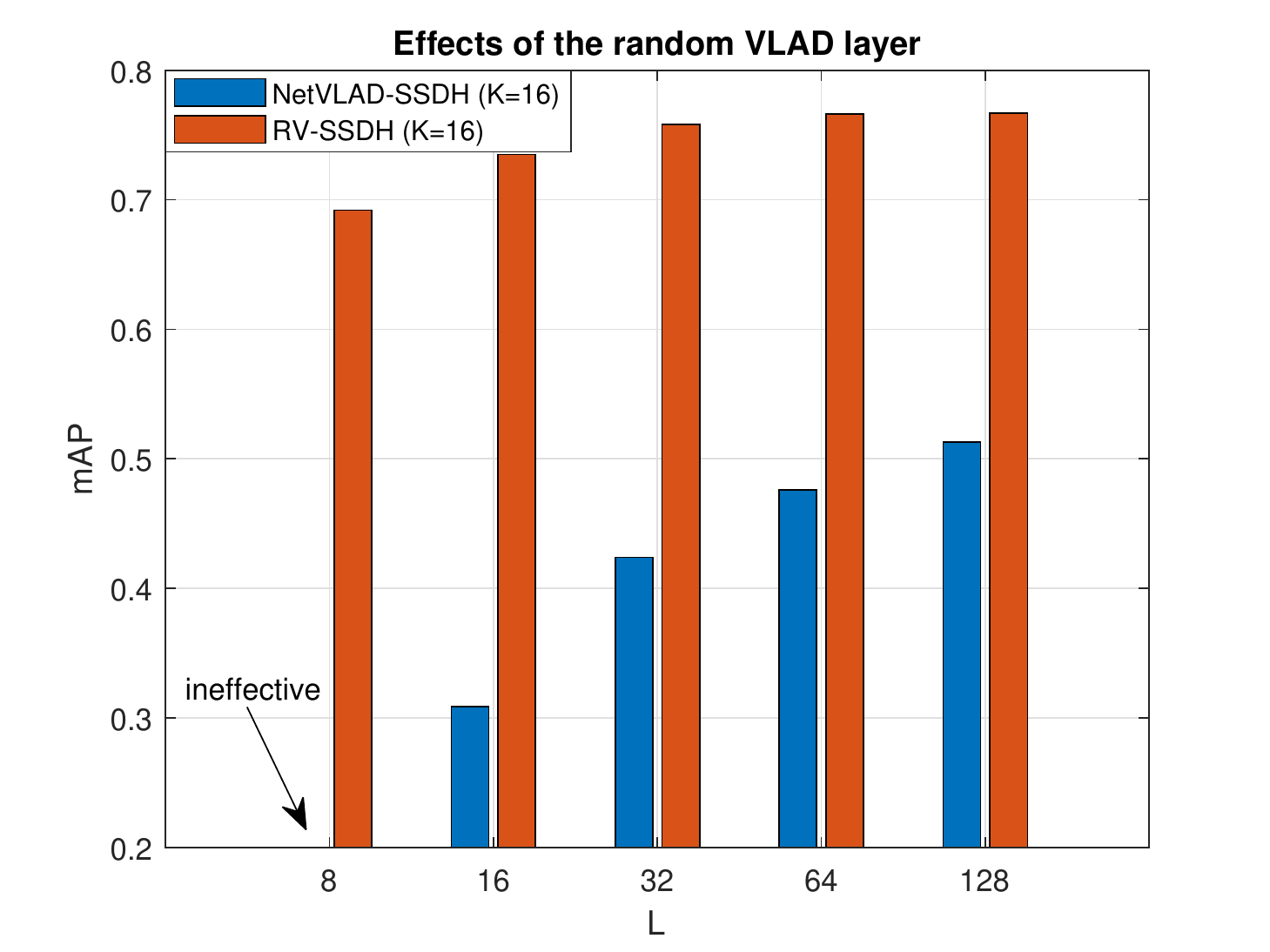}
\caption{mAP comparison (NetVLAD vs. Random VLAD, K=16).}
\label{fig_netvlad_vs_randomvlad_k16}
\vspace{-3mm}
\end{figure}

\begin{figure}
\centering
\includegraphics[scale=0.5]{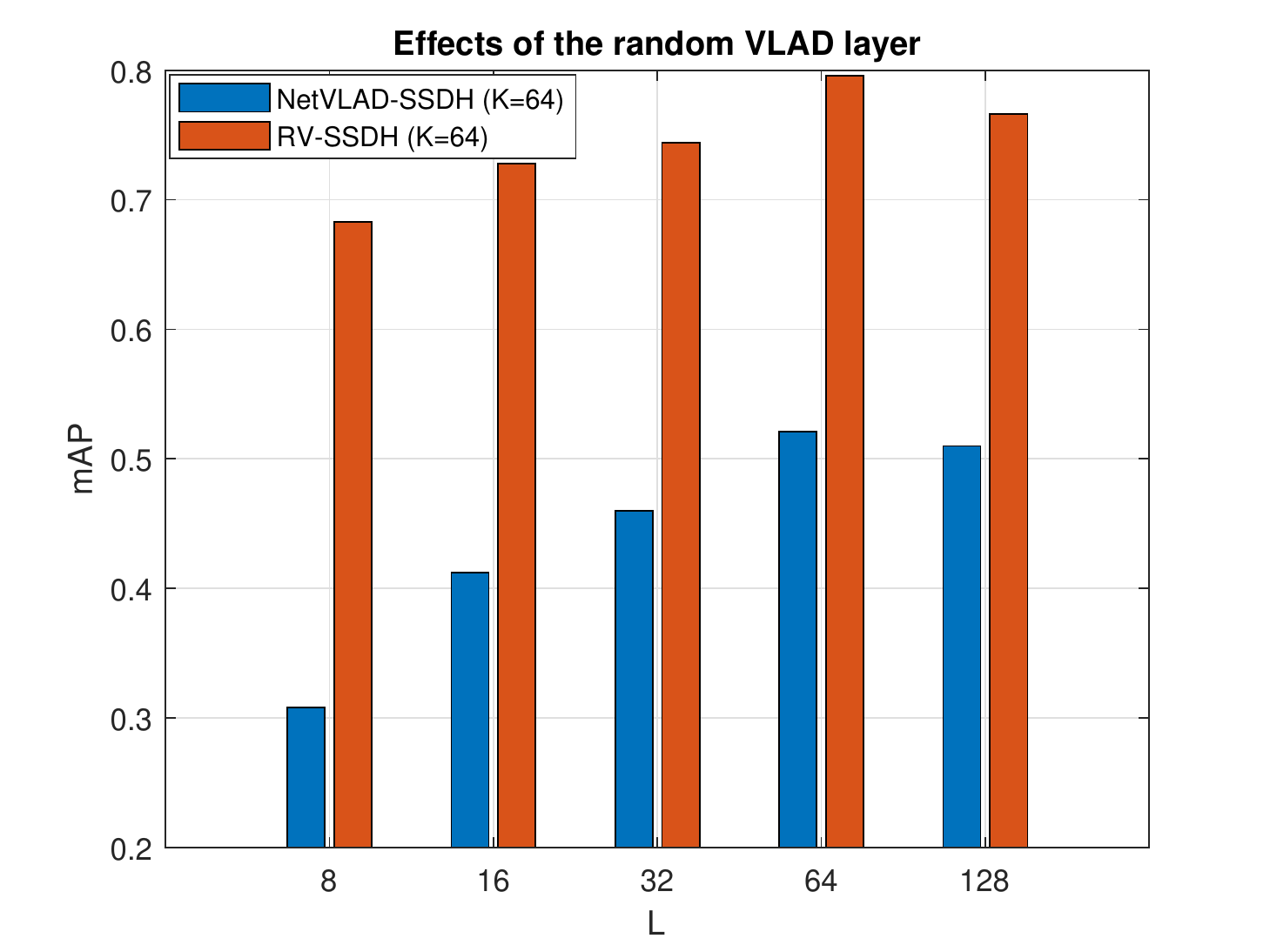}
\caption{mAP comparison (NetVLAD vs. Random VLAD, K=64).}
\label{fig_netvlad_vs_randomvlad_k64}
\vspace{-3mm}
\end{figure}

The random VLAD layer in RV-SSDH is inspired by VLAD and NetVLAD, but the random nature makes it different from the ancestors. What happens if the original NetVLAD is used in RV-SSDH? Some tests are performed to answer this question. The results are shown in Figures~\ref{fig_netvlad_vs_randomvlad_k16}--\ref{fig_netvlad_vs_randomvlad_k64}, where the dataset is CIFAR-10 and the base network is AlexNet. There is a significant performance drop if random VLAD is replaced by NetVLAD, and even a non-convergence (K=16, L=8). Therefore, ``NetVLAD+SSDH'' is not a good option.

\subsection{The effects of the transform layer}
\label{sect_experiment_transform_layer}
\begin{figure}
\centering
\includegraphics[scale=0.5]{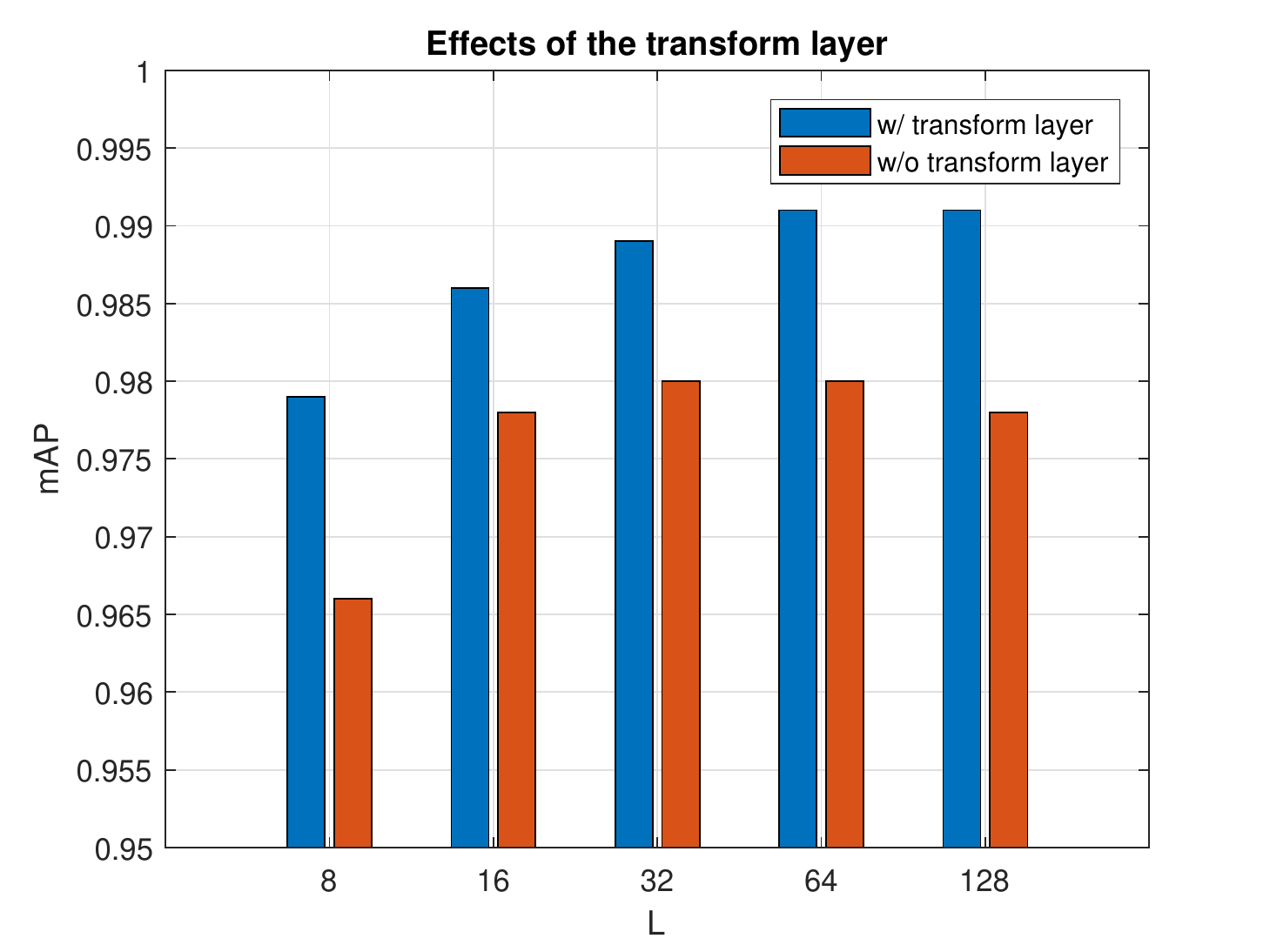}
\caption{Effects of transform layer (MNIST, ToyNet, RV-SSDH, K=16).}
\label{fig_effects_transform_mnist_toynet}
\end{figure}

\begin{figure}
\centering
\includegraphics[scale=0.5]{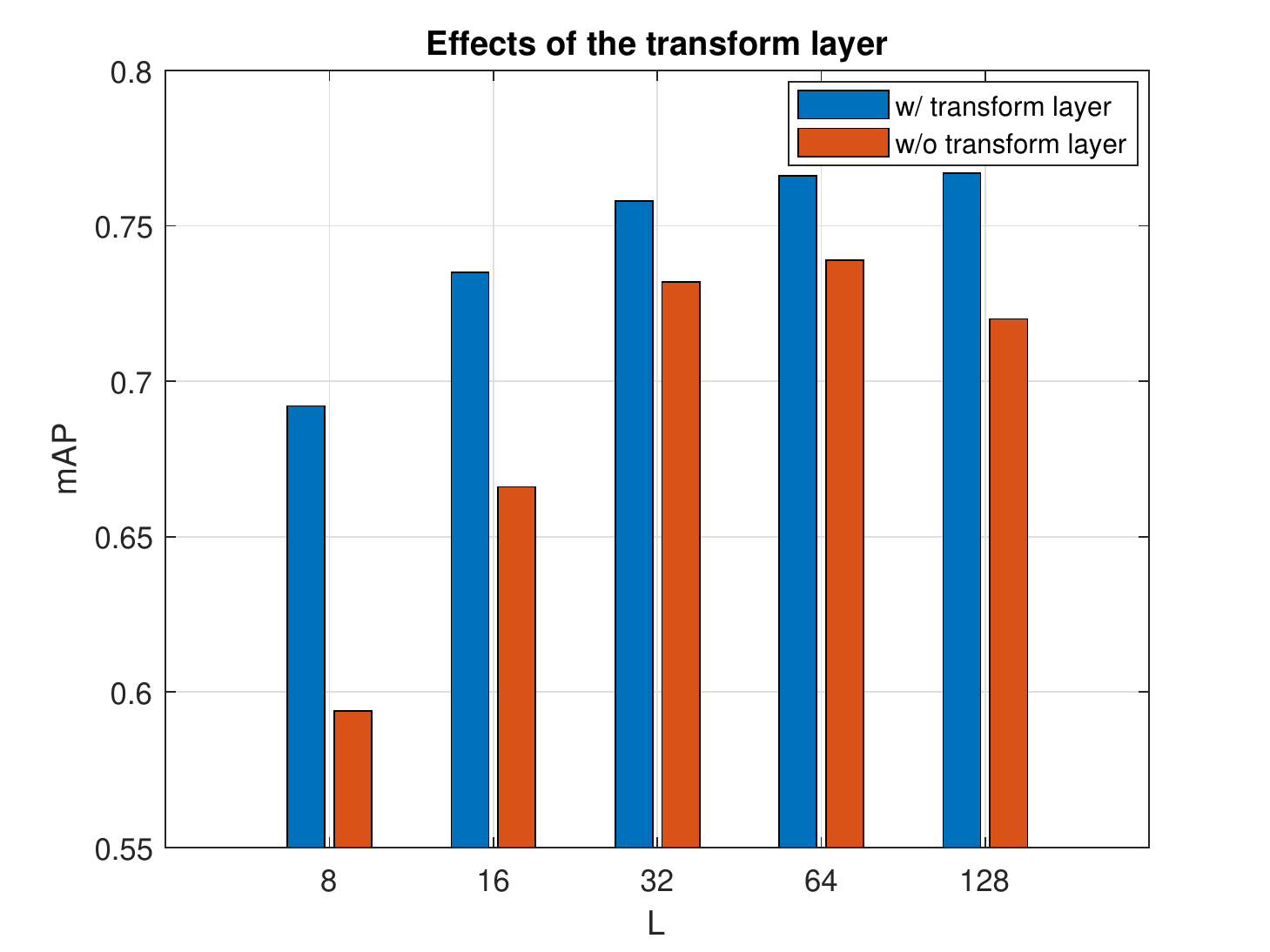}
\caption{Effects of transform layer (CIFAR-10, AlexNet, RV-SSDH, K=16).}
\label{fig_effects_transform_cifar10_alexnet}
\end{figure}

The transform layer is placed in between the random VLAD layer and the hash layer. The effects are verified by running another set of tests without the transform layer and comparing the mAP values. The results are shown in Fig.~\ref{fig_effects_transform_mnist_toynet}--\ref{fig_effects_transform_cifar10_alexnet} for two scenarios. In the best case, the transform layer increases mAP by approximately 10\%; in general, the gain is about 1\%--7\%. Therefore, it is good practice to keep the transform layer.

\subsection{Comparison with other hash algorithms}
Previous results already show the superiority of RV-SSDH over SSDH and NetVLAD. In order to know the position of RV-SSDH among other state-of-the-art hash algorithms, more tests are carried out with CIFAR-10 and VGG-F, in two settings with small hash lengths: Case-1 and Case-2 (see the description in \cite{Cakir2019}). The results are shown in Table~\ref{tab_map_comparison_case1}--\ref{tab_map_comparison_case2} and compared with some data collected from \cite{Cakir2019}. One can see that RV-SSDH still has advantages over a few baselines (such as the classic SH~\cite{Weiss2009}, ITQ~\cite{Gong2013}, and KSH~\cite{Liu2012}), but it is outperformed by MIHash~\cite{Cakir2019}, a recent deep hash algorithm. This is not a surprising result, because RV-SSDH uses point-wise training with complexity $O(N)$, while MIHash (and many other baselines) uses pairwise training with complexity $O(N^2)$. In fact, Case-1 is a disadvantageous situation for point-wise algorithms, where the training set is only 10\% of the total. According to the reasonable performance in Case-2, RV-SSDH is still an attractive ``economic'' solution, taking into account the low computational cost. 
Note that RV-SSDH can be modified to use pairwise or triplet training too, which is a promising path for future research; but on the other hand, pairwise training might prohibits large hash lengths due to the exponentially increasing complexity, thus reduces versatility.
\begin{table}
\centering
\caption{mAP comparison (CIFAR-10, VGG-F, Case 1).}
\begin{tabular}{l|llll}
\hline
\multirow{2}{*}{Method}	& \multicolumn{4}{c}{mAP vs. hash length}	\\
	\cline{2-5}		
	& 12 bits&	24 bits	& 32 bits & 48 bits \\
\hline
SH~\cite{Weiss2009}	& 0.183	& 0.164	&0.161	&0.161 \\
ITQ~\cite{Gong2013}	&0.237	& 0.246	&0.255	&0.261 \\
SPLH~\cite{Wang2010a} &	0.299	&0.33	&0.335	&0.33 \\

KSH~\cite{Liu2012} & 0.488 & 0.539 & 0.548 & 0.563\\
SDH~\cite{Shen2015} & 0.478 & 0.557 & 0.584 & 0.592\\
RV-SSDH & \bf{0.551} & \bf{0.589} & \bf{0.603} & \bf{0.598}\\
MIHash~\cite{Cakir2019} & 0.738  &0.775  &0.791  &0.816\\ 
\hline
\end{tabular}
\label{tab_map_comparison_case1}
\vspace{-3mm}
\end{table}

\begin{table}
\centering
\caption{mAP comparison (CIFAR-10, VGG-F, Case 2).}
\begin{tabular}{l|llll}
\hline
\multirow{2}{*}{Method}	& \multicolumn{4}{c}{mAP vs. hash length}	\\
	\cline{2-5}		
	& 12 bits&	24 bits	& 32 bits & 48 bits \\
\hline
DRSH~\cite{FangZhao2015}	&0.608	&0.611	&0.617	&0.618 \\
DRSCH~\cite{Zhang2015}	&0.615	&0.622	&0.629	&0.631 \\
RV-SSDH & \bf{0.899} & \bf{0.906} & \bf{0.908} & \bf{0.905}\\
DPSH~\cite{Li2016} & 0.903 & 0.885 & 0.915 & 0.911\\
MIHash~\cite{Cakir2019}  &0.927  &0.938  &0.942  &0.943 \\
\hline
\end{tabular}
\label{tab_map_comparison_case2}
\vspace{-3mm}
\end{table}

\section{Conclusion and discussion}
\label{sect_conclusion}
In this work, we propose RV-SSDH, a deep hash algorithm that incorporates VLAD (vector of locally aggreggated descriptors) into a neural network architecture. The core of RV-SSDH is a random VLAD layer coupled with a latent hash layer through a transform layer. It is a point-wise algorithm that can be efficiently trained by minimizing classification error and quantization loss. This novel construction significantly outperforms baselines such as NetVLAD and SSDH in both accuracy and complexity, thus offers an alternative trade-off in the state-of-the-art. Our future work might include pairwise or triplet training, adding GAN, or multiscale extension~\cite{Shi2018}.

Our experiment results also reveal some drawbacks of NetVLAD:1) the normalization steps are slow; 2) the initialization of anchors is cumbersome and inflexible (even ineffective); 3) it is not suitable for point-wise training. These issues make our random VLAD an interesting alternative.

\ifCLASSOPTIONcaptionsoff
  \newpage
\fi



\bibliographystyle{IEEEtran}
\bibliography{hashing}
%
%
%

%

\begin{IEEEbiographynophoto}{Li Weng} is currently an Assistant Professor at Hangzhou Dianzi University. He received his PhD in electrical engineering from University of Leuven (Belgium) in 2012. He worked on encryption, authentication, and hash algorithms for multimedia data. He then worked at University of Geneva (Switzerland) and Inria (France) on large-scale CBIR systems with emphasis on privacy protection. He was a post-doctoral researcher at IGN - French Mapping Agency. His research interests include multimedia signal processing, machine learning, and information security.
\end{IEEEbiographynophoto}

\begin{IEEEbiographynophoto}{Lingzhi Ye} received her Bachelor's degree in Automation from Hangzhou Dianzi University, Hangzhou, China, in 2017. She is currently working towards the Master’s degree at the Hangzhou Dianzi University.
Her research interests include deep learning, computer vision.
\end{IEEEbiographynophoto}

\begin{IEEEbiographynophoto}{Jiangmin Tian} received her B.S. degree in software engineering and PhD degree in control science and engineering from Huazhong University of Science and Technology, Wuhan, China, in 2010 and 2019 respectively. She is currently a lecturer at Hangzhou Dianzi University. Her research interests include machine learning and computer vision.
\end{IEEEbiographynophoto}

\begin{IEEEbiographynophoto}{Jiuwen Cao} received the B.Sc. and M.Sc. degrees from the School of Applied Mathematics, University of Electronic Science and Technology of China, Chengdu, China, in 2005 and 2008, respectively, and the Ph.D. degree from the School of Electrical and Electronic Engineering, Nanyang Technological University (NTU), Singapore, in 2013. From 2012 to 2013, he was a Research Fellow with NTU. Now, He is a Professor of Hangzhou Dianzi University, Hangzhou, China. His research interests include machine learning, artificial neural networks, intelligent data processing, and array signal processing.

He is an Associate Editor of \emph{IEEE Transactions on Circuits and Systems I: Regular paper}, \emph{Journal of the Franklin Institute}, \emph{Multidimensional Systems and Signal Processing}, and \emph{Memetic computing}. He has served as a Guest Editor of \emph{Journal of the Franklin Institute} and \emph{Multidimensional Systems and Signal Processing}.
\end{IEEEbiographynophoto}

\begin{IEEEbiographynophoto}{Jianzhong Wang} received the Bachelor's degree with the School of Computer Science and Engineering, XiDian University, Xi'an, China, in 1985, and the Master's degree with the School of Computer Science and Engineering, Zhejiang University, Hangzhou, China, in 1993, respectively.
He has been a Faculty with the Hangzhou Dianzi University, Hanzghou, since 1985, where he is currently a Professor. He served as the Vice Dean from 2000 and has been serving as the Dean since 2016 with the School of Automation. He has published extensively in international journals and conferences and authorized over 30 patents. His current research interests include computer information system development, computer control, embedded system, and system modeling and optimization. He was a recipient of a number of national science and technology awards, including the Second Prize of State-Level Teaching Award.
\end{IEEEbiographynophoto}







\end{document}